\documentclass[runningheads]{llncs}

\usepackage[final,year=2026,ID=xxxxx]{eccv}

\definecolor{cvprblue}{rgb}{0.21,0.49,0.74}
\usepackage[breaklinks,colorlinks,allcolors=cvprblue]{hyperref}
\usepackage{eccvabbrv}
\usepackage{graphicx}
\usepackage{booktabs}
\usepackage[accsupp]{axessibility}  
\usepackage{url}
\usepackage{enumitem}
\usepackage{subcaption}
\usepackage{graphicx}
\usepackage{multirow}
\usepackage{tabularx}
\usepackage{multirow}
\usepackage{booktabs} 
\usepackage{caption}  
\usepackage{float}
\usepackage{afterpage}
\usepackage{colortbl}
\usepackage{tabulary,multirow,overpic,xcolor}
\usepackage{makecell}
\usepackage{algorithm} 
\usepackage{algpseudocode} 
\usepackage{siunitx}  
\usepackage{amsmath}
\usepackage{mathtools} 

\usepackage{orcidlink}
\usepackage{bbding}

\def\eg{\emph{e.g., }}

\def\name{MixGRPO}
\def\bfx{\mathbf{x}}

\begin{document}

\title{\name{}: Unlocking Flow-based GRPO Efficiency with Mixed ODE-SDE}


\author{Junzhe Li\inst{1,2,3}\thanks{Equal contribution.~(lijunzhe1028@gmail.com)} \and
Yutao Cui\inst{1}\textsuperscript{*} \and
Tao Huang\inst{1}\textsuperscript{*} \and
Weijie Kong\inst{1} \and
Chuxuan Zeng\inst{4} \and
Yiming Cheng\inst{5} \and
Yinping Ma\inst{3} \and
Chun Fan\inst{3} \and
Miles Yang\inst{1} \and
Zhao Zhong\inst{1}\Envelope \and
Liefeng Bo\inst{1}}

\authorrunning{J.~Li et al.}

\institute{Hunyuan, Tencent, China \and
School of Computer Science, Peking University, China \and
Computer Center, Peking University, China \and
China Mobile Communications Group, China \and
Department of Engineering Physics, Tsinghua University, China}

\maketitle
\begingroup
\renewcommand{\thefootnote}{\Envelope}
\footnotetext{Corresponding authors.}
\endgroup

\begin{abstract}
Although GRPO substantially improves flow-matching models for human preference alignment in vision generation, mainstream methods such as DanceGRPO rely on Global-Stochastic Differential Equations (SDE) sampling across full timesteps in the Markov Decision Process (MDP), which remains computationally inefficient. In this paper, we propose \textit{\textbf{\name{}}}, a novel framework that leverages the flexibility of mixed sampling strategies through the integration of SDE and Ordinary Differential Equations (ODE). This redesign streamlines optimization within the MDP, \textbf{delivering gains in both generation performance and training efficiency}. Specifically, \name{} introduces a sliding window mechanism, using SDE sampling and GRPO-guided optimization only within the window, while applying ODE sampling outside. This design confines sampling randomness to the time-steps within the window, thereby reducing the optimization overhead, and allowing for more focused gradient updates to accelerate convergence. Additionally, as time-steps beyond the sliding window are not involved in optimization, higher-order solvers are supported for faster sampling. So we present a faster variant, termed \textit{\textbf{\name{}-Flash}}, which further improves training efficiency while achieving comparable performance. MixGRPO exhibits substantial gains across multiple dimensions of human preference alignment, outperforming DanceGRPO in both effectiveness and efficiency, \textbf{with nearly 50\% lower training time}. Notably, MixGRPO-Flash further \textbf{reduces training time by 71\%}.\footnote{Code is available in https://github.com/Tencent-Hunyuan/MixGRPO}
\end{abstract}  
\section{Introduction}
\label{sec:intro}

Recent advances~\cite{liu2025flow, xue2025dancegrpo, liu2025improving, xu2023imagereward} in vision generation have demonstrated that probability flow models can achieve improved performance by incorporating Reinforcement Learning from Human Feedback (RLHF)~\cite{ouyang2022training} strategies during the post-training stage to maximize rewards. Specifically, methods~\cite{liu2025flow, xue2025dancegrpo} based on Group Relative Policy Optimization (GRPO)~\cite{shao2024deepseekmath}, have recently been studied, achieving optimal alignment with human preferences. 

\begin{figure}[!t]
    \centering
    \resizebox{0.88\linewidth}{!}{%
        \includegraphics{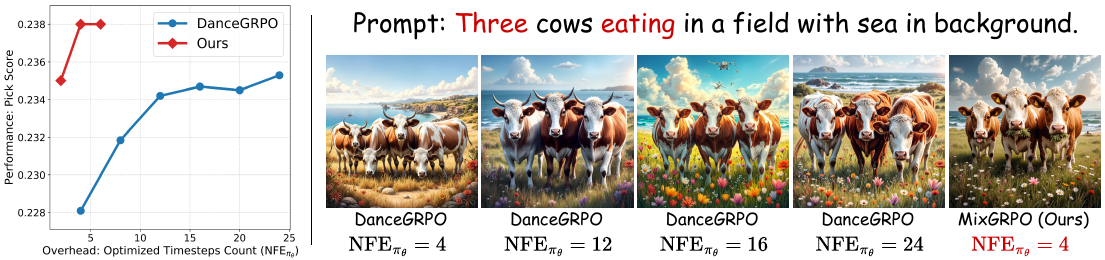}%
    }
    \vspace{-0.8em}
    \caption{Comparison of MixGRPO and DanceGRPO with {\bf varying denoising timesteps optimized}. MixGRPO achieves higher performance with lower overhead.}
    \label{fig:prelab_degradation}
    \vspace{-1.5em}
\end{figure}

Current GRPO methods in probability flow models, such as Flow-GRPO and DanceGRPO, combine a rollout phase followed by an online optimization phase. During rollouts, they employ Stochastic Differential Equation (SDE) sampling at each denoising step to introduce randomness, thereby enabling the stochastic exploration required by RLHF. This allows the entire denoising process to be framed as a Markov Decision Process (MDP) in a stochastic environment, where GRPO is then applied to optimize the complete state-action sequence in the online optimization phase.
However, the requirement to optimize the entire denoising steps introduces two challenges, \emph{prohibitive computational overhead} and \emph{unfocused and inefficient optimization}. \romannumeral 1) The overhead arises because computing the policy ratio requires separate full-trajectory sampling from both the old ($\pi_{\theta_\text{old}}$) and new policies ($\pi_\theta$). Although DanceGRPO proposes a workaround by optimizing a random subset of steps, our findings in Figure~\ref{fig:prelab_degradation} reveal a critical trade-off: reducing the subset size to save computation severely compromises model performance.
\romannumeral 2) Early-step gradients prioritize global structure while late-step gradients focus on fine details, leading to conflicting update signals. This may result in an inefficient training process.

To address these issues, we propose \name{}, a method that optimizes a strategic subset of denoising steps to drastically reduce computational overhead while ensuring a more focused and efficient optimization.
\afterpage{\afterpage{%
\begin{figure*}[!t]
    \centering
    \resizebox{0.9\linewidth}{!}{%
        \includegraphics{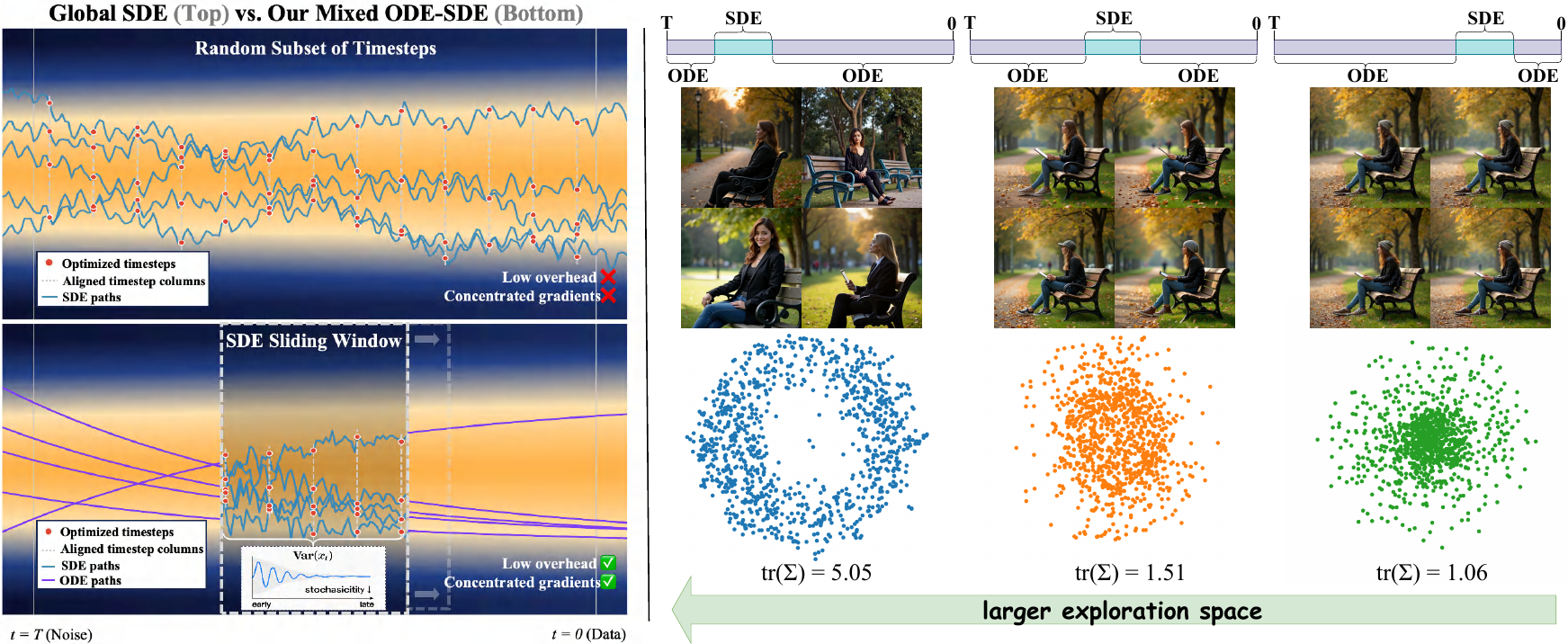}%
    }
    \vspace{-0.8em}
    \caption{(Left) \textbf{Methods Comparison.} By employing the mixed ODE-SDE and sliding window strategy, MixGRPO reduces the number of optimized timesteps and achieves more efficient training than DanceGRPO~\cite{xue2025dancegrpo}. (Right) \textbf{SDE Exploration Analysis.} As training progresses, MixGRPO gradually slides the SDE window from a high-noise to a low-noise regime, causing the exploration space to shrink and the distribution of sampled images to transition from dispersed to concentrated.}
    \label{fig:prelab_tsne}
    \vspace{-1.5em}
\end{figure*}%
}}
The policy optimization is confined only to the SDE sub-interval. This interval operates as a sliding window, systematically advancing from low-SNR (signal-to-noise ratio) steps to high-SNR denoising steps as training progresses. This curriculum-based approach is conceptually analogous to temporal discounting in RL~\cite{pitis2019rethinking, amit2020discount, hu2022role}, as it prioritizes the optimization of initial, high-impact steps that establish global structure from a vast exploration space (Figure~\ref{fig:prelab_tsne} Right), before progressively shifting focus to the refinement of local details. This selective optimization not only reduces the computational burden but also fosters a more dedicated exploration. Furthermore, this decoupling allows us to accelerate the deterministic ODE portions with fast solvers (\eg{} DPMSolver++~\cite{lu2022dpm++}), since their posterior distributions are irrelevant to the optimization, thereby reducing rollout time without compromising final image quality. Finally, we adopt Coefficients-Preserving Sampling (CPS)~\cite{wang2025coefficients} as a stable stochastic sampler in our framework, which reduces sampling artifacts and yields more reliable reward feedback.

To evaluate the performance and efficiency of \name{}, we first conduct comprehensive experiments on FLUX~\cite{flux2024} by directly comparing against DanceGRPO~\cite{xue2025dancegrpo}, a representative Global-SDE baseline. Results show that our method consistently surpasses DanceGRPO in both effectiveness and efficiency. In particular, \name{} improves ImageReward~\cite{xu2023imagereward} from 1.088 to 1.629, exceeding DanceGRPO's 1.436, while generating images with better semantic fidelity, aesthetics, and fewer distortions. Moreover, compared with DanceGRPO under its official setting, \name{} reduces training time by nearly 50\%, and \name{}-Flash further achieves a 71\% training-time reduction. These results highlight the advantage of our Mixed ODE-SDE with sliding-window formulation. We further validate \name{} on both in-domain and out-of-domain reward models, and extend evaluation to cross-dataset settings, where \name{} demonstrates strong generalization. Beyond this, we verify robustness across different backbone models and scenarios, including SD3.5-M~\cite{esser2024scaling} and HunyuanImage-3.0~\cite{cao2025hunyuanimage}, and further extend the method to video generation on HunyuanVideo-1.5~\cite{hunyuanvideo2025}, where consistent gains are observed. Finally, to demonstrate the competitiveness of \name{}, we compare with other RL-based alignment methods, including Flow-DPO~\cite{liu2025improving}, Flow-GRPO~\cite{liu2025flow}, and DiffusionNFT~\cite{zheng2025diffusionnft}. Across these comparisons, \name{} achieves advanced alignment performance while delivering substantial training acceleration.


To summarize, the key contributions are outlined below:
\begin{itemize} [leftmargin=*]
\item We propose a mixed ODE-SDE framework for GRPO training with flow-based models. This approach alleviates computational overhead by confining the stochastic optimization to a specific sub-interval of the MDP.

\item We introduce an SDE sliding window to dynamically schedule the optimized timesteps. By guiding the optimization from broad exploration to fine-grained refinement, this strategy significantly enhances performance.

\item Our hybrid framework enables the use of high-order ODE solvers to accelerate the deterministic sampling of $\pi_{\theta_\text{old}}$ during GRPO training. This yields substantial speedups with negligible performance degradation.

\item Comprehensive experiments across multiple frameworks demonstrate the versatility of \name{}. Our method consistently achieves substantial performance gains while significantly reducing training overhead, highlighting its broad applicability and effectiveness.

\end{itemize}
\section{Related Work}

RLHF for vision generation has advanced from differentiable reward-based fine tuning~\cite{xu2023imagereward} to PPO-style~\cite{miao2024training} and DPO-style~\cite{liu2025improving}, and more recently to GRPO-based optimization in flow-based models, where SDE-induced stochasticity enables broader exploration. Methods such as Flow-GRPO and DanceGRPO improve alignment quality, but still suffer from a key bottleneck: optimizing many imbalanced timesteps, which leads to redundant effective MDP horizons under Global-SDE sampling. In parallel, sampling research has progressed from ODE solvers to high-order variants. Motivated by the SDE-ODE equivalence in probability flow, we establish a mixed SDE-ODE MDP formulation and systematically explore different scheduling strategies for optimizing selected SDE timesteps, establishing a more efficient paradigm, \name{}. A more comprehensive review of RL for Vision Generation and Sampling Methods is provided in \textbf{Appendix~\ref{app: related_work}}.
\section{Method}
\subsection{Mixed ODE-SDE Sampling in GRPO}
\label{sec: mixed ODE-SDE}
According to Flow-GRPO~\cite{liu2025flow}, the SDE sampling in flow matching can be framed as a Markov Decision Process (MDP) \(\left( \mathcal{S}, \mathcal{A}, \rho_0, P, \mathcal{R} \right)\). The agent produces a trajectory during the discrete sampling process, defined as $\Gamma = \{(\mathbf{s}_t, \mathbf{a}_t)\}_{t=0}^{T}$, where the reward is provided only at the final step by the reward model, specifically $\mathcal{R}(\mathbf{s}_i, \mathbf{a}_i) \triangleq R(\mathbf{x}_T, c)$ if $i=T$, and 0 otherwise.

In \name{}, we propose a sampling method that combines SDE and ODE. \name{} defines a time interval $S=[t_l, t_r) \subseteq [0, 1)$, which corresponds to a subset of denoising timesteps, such that $0 \le l < r \le T $ and $t_i=\frac{i}{T}$. We use SDE sampling within the interval \( S \) and ODE sampling outside, while \( S \) shifts along the denoising direction throughout the training process (See Figure~\ref{fig:prelab_tsne} Left).
\name{} restricts the stochastic exploration to the interval \( S \), shortening the sequence length of the MDP to a subset $\Gamma_{\text{\name{}}} = \{(\mathbf{s}_t, \mathbf{a}_t)\}_{t=l}^{r}$ and requires policy optimization only within \( S \):
{\small
\begin{equation}
\max_{\theta} \mathbb{E}_{\Gamma_\text{MixGRPO}\sim\pi_\theta} \left[ \sum_{t\in S} \left( \mathcal{R}(\mathbf{s}_t, \mathbf{a}_t) - \beta D_\text{KL}\big( \pi(\cdot \mid \mathbf{s}_t) \| \pi_\text{ref}(\cdot \mid \mathbf{s}_t) \big) \right) \right],
\label{eq:RL}
\end{equation}
}
MixGRPO reduces training overhead while also enabling more concentrated gradient updates. Next, we derive the specific sampling form. For a deterministic reverse \textit{probability flow ODE}~\cite{song2020score}, it takes the following form:
{\small
\begin{equation}
\label{eq: prob_ode}
\frac{\mathrm{d}\mathbf{x}_t}{\mathrm{d}t} = 
f(\mathbf{x}_t,t) - 
\frac{1}{2}g^2(t)\nabla_{\mathbf{x}_t}\log{q_t(\mathbf{x}_t)},
\end{equation}
}
where $q_t(\bfx_t)$ represents the evolution process of the reverse probability distribution from $0$ to $T$. $\nabla_{\mathbf{x}_t}\log{q_t(\mathbf{x}_t)}$ is the \textit{score function} at time $t$.
According to the Fokker-Planck equation~\cite{risken1996fokker, oksendal2003stochastic}, ~\cite{song2020score} has demonstrated that Eq. (\ref{eq: prob_ode}) has the following equivalent \textit{probability flow SDE}, which maintains the same marginal distribution at each time \( t \):
{\small
\begin{equation}
\label{eq: prob_sde}
\frac{\mathrm{d}\mathbf{x}_t}{\mathrm{d}t} = 
f(\mathbf{x}_t, t) - 
g^2(t)\nabla_{\mathbf{x}_t}\log{q_t(\mathbf{x}_t)} + g(t)\frac{\mathrm{d}\mathbf{w}}{\mathrm{d}t}.
\end{equation}
}

In \name{}, we mix ODE and SDE for sampling. Under standard regularity assumptions (continuous $g(t)$, locally Lipschitz drift/score fields, finite second moments, and sufficiently small $\Delta t$), the mixed process preserves the same time-marginal dynamics as the corresponding probability-flow ODE up to discretization/score-approximation error; a full proof is provided in Appendix~\ref{app: proof_convergence}. The specific form is as follows:
{\small
\begin{equation}
\label{eq: mix_sampling_short}
\mathrm{d}\mathbf{x}_t \! = \!
\begin{cases}
    \left[f(\mathbf{x}_t, t) \!  -  g^2(t)\mathbf{s}_t(\mathbf{x}_t)\right]\mathrm{d}t + g(t)\mathrm{d}\mathbf{w}, \! \! & \text{if}~t \in S, \\
    \left[f(\mathbf{x}_t, t) \! - \frac{1}{2}g^2(t)\mathbf{s}_t(\mathbf{x}_t)\right]\mathrm{d}t, \! \! & \text{otherwise},
\end{cases}
\end{equation}
}
where we define the score function as $\mathbf{s}_t\ \triangleq \nabla_{\mathbf{x}_t}\log{q_t(\mathbf{x}_t)}$.  In particular, for Flow Matching (FM)~\cite{lipman2022flow}, especially the Rectified Flow (RF)~\cite{liu2022flow}, the sampling process can be viewed as a deterministic ODE:
{\small
\begin{equation}
\label{eq: flow_matching}
\frac{\mathrm{d}\mathbf{x}_t}{\mathrm{d}t} = \mathbf{v}_t.
\end{equation}
}
Eq. (\ref{eq: flow_matching}) is actually a special case of the Eq. (\ref{eq: prob_ode}) with $\mathbf{v}_t=f(\mathbf{x}_t,t) - 
\frac{1}{2}g^2(t)\mathbf{s}_t(\mathbf{x}_t)$. So we can derive mixed ODE-SDE sampling form for RF as follows:
{\small
\begin{equation}
\begin{aligned}
\mathrm{d}\mathbf{x}_t \!=\!
\begin{cases}
\left[ \mathbf{v}_t \! - \!
\frac{1}{2}g^2(t)\mathbf{s}_t(\mathbf{x}_t)  \right] \mathrm{d}t + g(t)\mathrm{d}\mathbf{w},
&
\text{if}~t \in S,
\\
\mathbf{v}_t \mathrm{d}t,
&
\text{otherwise}.
\end{cases}
\end{aligned}
\end{equation}
}
In the RF framework, the model is used to predict the velocity field as $\mathbf{v}_\theta(\mathbf{x}_t, t) = \frac{\mathrm{d} \mathbf{x}_t}{\mathrm{d}t}$. Following~\cite{liu2025flow}, the \textit{score function} is represented as $\mathbf{s}_t(\mathbf{x}_t) = -\frac{\mathbf{x}_t}{t} - \frac{1-t}{t}\mathbf{v}_\theta(\mathbf{x}_t, t)$. The $g(t)$ is represented as the standard deviation of the noise $g(t)=\sigma_t$. According to the definition of the standard Wiener process, we use $\mathrm{d}\mathbf{w}=\sqrt{\mathrm{d}t} \epsilon$, where $\epsilon \sim \mathcal{N}(0, \mathbf{I})$. Applying Euler-Maruyama discretization for SDE and Euler discretization for ODE, we build the final denoising process in \name{}:
{\small
\begin{equation}
\label{eq: mix_sampling}
\mathbf{x}_{t+\Delta t} =
\begin{cases}
    \mathbf{x}_t + \boldsymbol{\mu}_\theta(\mathbf{x}_t, t)\Delta t + \sigma_t \sqrt{\Delta t}\epsilon, & \text{if } t \in S, \\
    \mathbf{x}_t + \mathbf{v}_\theta(\mathbf{x}_t, t) \Delta t, & \text{otherwise},
\end{cases}
\end{equation}
}
where the SDE drift term for the sampling process is defined as $\boldsymbol{\mu}_\theta(\mathbf{x}_t, t) \triangleq \mathbf{v}_\theta(\mathbf{x}_t, t) + \frac{{\sigma^2_t} \left( \mathbf{x}_t+ \left(1-t \right)\mathbf{v}_\theta(\mathbf{x}_t, t) \right)}{2t}$. For timesteps $t\in S$, Eq.~(\ref{eq: mix_sampling}) induces (via Euler--Maruyama) the conditional policy density
{\small
\begin{equation}
q_\theta(\mathbf{x}_{t+\Delta t}\mid\mathbf{x}_t,c)=\mathcal{N}\!\left(\mathbf{x}_t+\boldsymbol{\mu}_\theta(\mathbf{x}_t,t)\Delta t,\sigma_t^2\Delta t\,\mathbf{I}\right),\quad t\in S,
\end{equation}
}
with the same form for $q_{\theta_{\text{old}}}$. For $t\notin S$, the update is deterministic (Dirac transition), and these steps are excluded from GRPO ratio/KL evaluation. According to Eq.~(\ref{eq:RL}), GRPO optimization is only performed on timesteps within the interval \(S\) for each group of \(N\) samples. This design not only shortens the effective MDP horizon but also concentrates gradient updates, leading to the following training objective:
{\footnotesize
{\small
\begin{equation}
\begin{aligned}
\mathcal{J}_{\text{MixGRPO}}(\theta)=&\;\mathbb{E}_{c\sim\mathcal{C},~\{\mathbf{x}_T^i\}_{i=1}^{N}\sim\pi_{\theta_{\text{old}}}(\cdot\mid c)}\Bigg[
\frac{1}{N}\sum_{i=1}^{N}\frac{1}{|S|}\sum_{t\in S}
\Big(\min\big(r_t^i(\theta)A^i, \\
&\operatorname{clip}(r_t^i(\theta),1-\varepsilon,1+\varepsilon)A^i\big)-\beta\mathcal{J}_{\text{KL}}\Big)\Bigg],
\end{aligned}
\end{equation}
}
}
where $\varepsilon$ is the clipping threshold, $r^i_t(\theta)$ is the policy ratio, and $A^i$ is the advantage score. For optimized timesteps ($t\in S$), we parameterize the policy by the mean of the Gaussian transition above, while the covariance is fixed to $\sigma_t^2\Delta t\,\mathbf{I}$ for both $\theta$ and $\theta_{\text{old}}$. Therefore, the ratio and KL are defined as follows according to~\cite{liu2025flow}:
{\footnotesize
{\small
\begin{equation}
\begin{aligned}
    r^i_t(\theta) &= \frac{q_\theta(\mathbf{x}_{t+\Delta t}|\mathbf{x}_{t},c)}{q_{\theta_{\text{old}}}(\mathbf{x}_{t+\Delta t}|\mathbf{x}_{t},c)}, \quad t\in S,
    \quad
    A^i = \frac{R\left(\mathbf{x}^i_T,c \right) - \text{mean}\left( \{ R\left(\mathbf{x}^i_T,c \right) \}^{N}_{i=1} \right)}{\text{std}\left( \{ R\left(\mathbf{x}^i_T,c \right) \}^{N}_{i=1} \right)}, \\
    \mathcal{J}_\text{KL}\! &=\! D_\text{KL}\!\left(q_\theta(\cdot|\mathbf{x}_{t},c)\,||\,q_{\theta_\text{old}}(\cdot|\mathbf{x}_{t},c)\right) \!= \! \frac{||\mathbf{x}_{t+\Delta t}(\theta)\! - \!\mathbf{x}_{t+\Delta t}(\theta_\text{old})||^2}{2\sigma_t^2 \Delta t}, \quad t\in S.
\end{aligned}
\end{equation}
}
}
\name{} reduces the NFE of $\pi_\theta$ and optimized timesteps compared to Global-SDE sampling. However, the NFE of $\pi_{\theta_\text{old}}$ is not reduced, as complete inference is required to obtain the final image for reward calculation. In Section~\ref{sec:high_order_solver}, we will introduce the use of higher-order ODE solvers, which also reduce the NFE of $\pi_{\theta_\text{old}}$ leading to further speedup. In summary, the mixed ODE-SDE sampling significantly streamlines the MDP, enhancing training efficiency by lowering optimization overhead and enabling more focused gradient updates.

\begin{algorithm}[t]
\caption{\name{} Training Process}
\label{algo:mixgrpo}
\small
\begin{algorithmic}[1]
\Require initial policy model $\pi_\theta$; reward models $\{R_k\}_{k=1}^K$; prompt dataset $\mathcal{C}$; total sampling steps $T$; number of samples per prompt $N$; 
\Require sliding window $W(l)$, window size $w$, shift interval $\tau$, window stride $s$
\State Init left boundary of $W(l)$: $l \gets 0$
\For{training iteration $m=1$ \textbf{to} $M$}
    \State Sample batch prompts $\mathcal{C}_b \sim \mathcal{C}$
    \State Update old policy model: $\pi_{\theta_{\text{old}}} \gets \pi_\theta$
    \For{each prompt $\mathbf{c} \in \mathcal{C}_b$}
        \State Init the same noise $\mathbf{x_0} \sim \mathcal{N}(0,\mathbf{I})$ \Comment{according to DanceGRPO~\cite{xue2025dancegrpo}}
        \For{generate $i$-th image from $i=1$ \textbf{to} N}
            \For{sampling timestep $t=0$ \textbf{to} $T-1$} 
                \If{$t \in W(l)$}
                    \State Use SDE Sampling to get $\mathbf{x}^{i}_{t+1}$ 
                \Else
                    \State Use ODE Sampling to get $\mathbf{x}^{i}_{t+1}$
                \EndIf
            \EndFor

        \EndFor
        \For{$i$-th image from $i=1$ \textbf{to} N}
            \State Calculate advantages: 
            $A_i \gets \sum_{k=1}^K \frac{R(\mathbf{x}^i_T, \mathbf{c})^i_k - \mu_k}{\sigma_k}$ 
        \EndFor
        \For{optimization timestep $t \in W(l)$} 
            \State Update policy model: $\theta \gets \theta + \eta\nabla_\theta\mathcal{J}$
        \EndFor
    \EndFor
    \If{$m \bmod \tau~\text{is}~0$} \Comment{move sliding window}
        \State $l \gets \min(l+s,~T-w)$ 
    \EndIf
\EndFor
\end{algorithmic}
\end{algorithm}

\subsection{Sliding Window as Optimized Timestep Scheduler}
\label{sec: sliding_window_method}

\begin{figure}[t]
    \centering
    \resizebox{0.88\linewidth}{!}{%
      \includegraphics{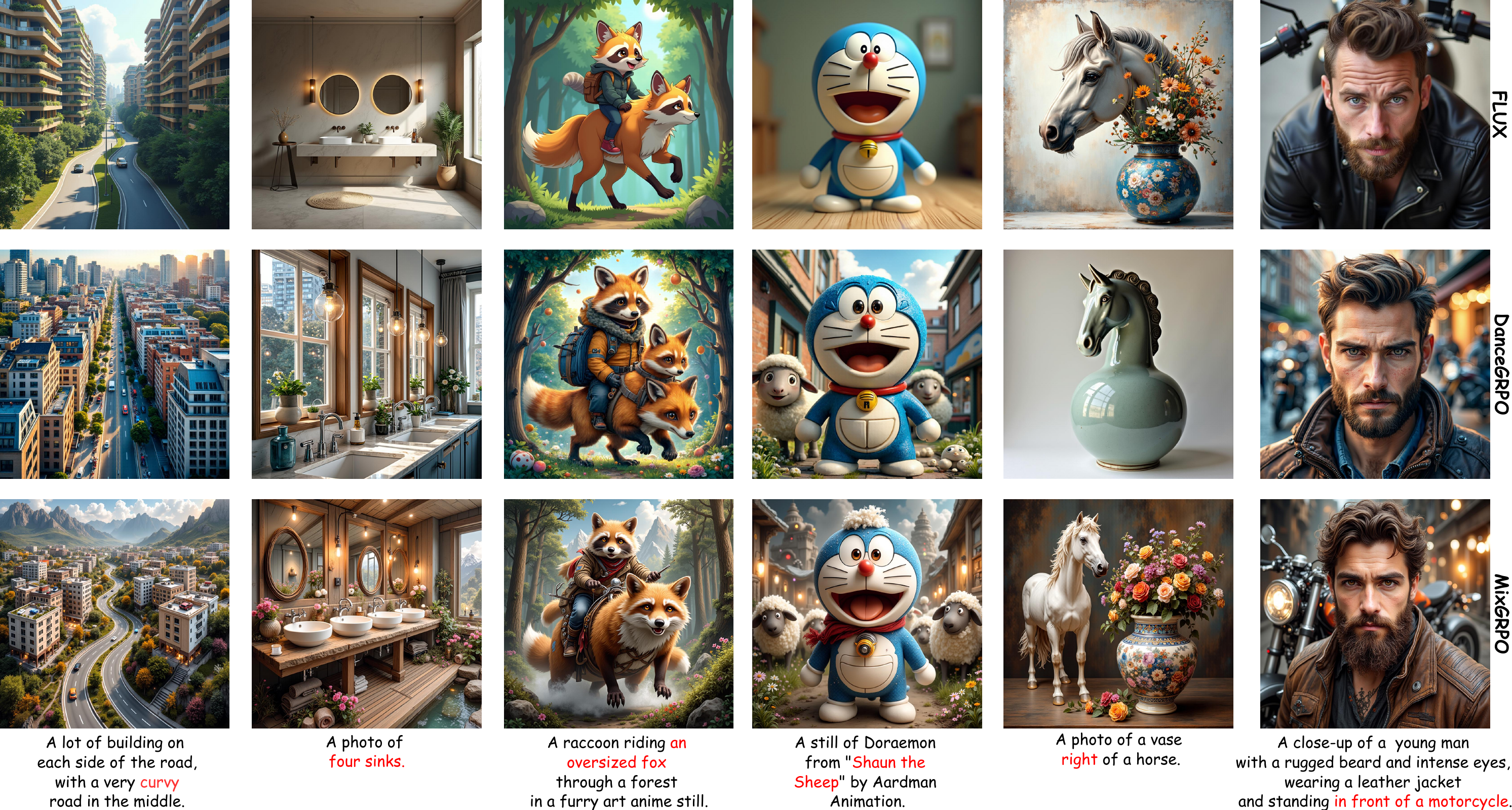}%
    }
    \vspace{-0.8em}
    \caption{Qualitative comparison. \name{} achieve superior performance in semantics, aesthetics and text-image alignment.}
    \label{fig:multi_reward_performance_comparation}
    \vspace{-1.5em}
\end{figure}

In this section, we will introduce the sliding window to describe the movement of $S$, which leads to a significant improvement in the quality of the generated images. Along the denosing time-steps $\{0, 1,\ldots,T-1\}$, \name{} defines a SDE sliding window $W(l)$ and optimization is only employed at the timesteps within $W(l)$.
\begin{equation}
    W(l) = \{t_l, t_{l+1}, \dots,t_{l+w-1}\}, \quad l \leq T-w,
\end{equation}
where $l$ is the \textit{left boundary} of the sliding window. The movement strategy during training is governed by three hyperparameters: \textbf{(1) window size $w$}, which determines the number of timesteps included in each window; \textbf{(2) shift interval $\tau$}, which specifies the number of training steps between two consecutive window shifts; and \textbf{(3) window stride $s$}, which indicates how many denoising timesteps the left boundary $l$ advances at each shift. In this work, we use a fixed progressive-constant scheduler as a simple and stable default across experiments. The detailed sliding-window strategy and the \name{} algorithm are provided in Algorithm~\ref{algo:mixgrpo}.

In MixGRPO, denoising follows a natural stochastic-to-deterministic trajectory along probability flow. We exploit this property with a sliding-window curriculum: optimization starts in low-SNR (high-stochasticity) regions and progressively shifts to high-SNR (more deterministic) regions for refinement, as illustrated in Figure~\ref{fig:prelab_tsne} (right). This coarse-to-fine schedule stabilizes early optimization and improves final alignment quality. Table~\ref{tab:abl_SW_Strategy} shows that progressive movement consistently outperforms random selection, while even a frozen window on early timesteps remains competitive, indicating that early denoising steps contribute disproportionately to optimization payoff~\cite{karras2022elucidating, esser2024scaling}. 


\subsection{Trade-off Between Overhead and Performance}
\label{sec:high_order_solver}
Unlike DanceGRPO~\cite{xue2025dancegrpo}, which relies on Global-SDE sampling, \name{} employs a mixed ODE-SDE method, allowing the use of higher-order ODE solvers to accelerate GRPO training-time sampling. However, accelerating the ODE phase before the SDE window amplifies the solver's numerical errors due to the window's stochasticity, severely degrading image quality and corrupting the reward signal (See Appendix~\ref{app: before_ODE_error} for details). In contrast, accelerating the ODE phase after the window provides a significant speed-up without compromising the image fidelity required for reliable reward evaluation. Therefore, \name{} exclusively accelerates the post-window ODE timesteps.

\cite{gao2025diffusionmeetsflow} has demonstrated the equivalence between the ODE sampling of flow matching models (FM) and DDIM, and Section~\ref{sec: mixed ODE-SDE} has also shown that diffusion probabilistic models (DPM) and FM share the same ODE form during the denoising process. Therefore, the higher-order ODE solvers \eg{} DPM-Solver Series~\cite{lu2022dpm, lu2022dpm++, zheng2023dpm}, UniPC~\cite{zhao2023unipc} designed for DPM sampling acceleration are also applicable to FM. We have reformulated DPM-Solver++~\cite{lu2022dpm++} to apply it in the FM framework for ODE sampling acceleration and released detailed derivations in Appendix~\ref{app: dpm_for_rf}.

By applying higher-order solvers, we achieve acceleration in the sampling of \( \pi_{\theta_{\text{old}}} \) during GRPO training, which is essentially a balance between overhead and performance. Excessive acceleration leads to fewer timesteps, which inevitably results in a decline in image generation quality, thereby accumulating errors in the computation of rewards. We have found in practice that the 2nd-order DPM-Solver++ is sufficient to provide significant acceleration while ensuring that the generated images align well with human preferences in Table \ref{tab:abl_Solver_order}. 

Ultimately, we integrate DPM-Solver++ with both progressive and frozen sliding-window strategies, leading to MixGRPO-Flash and MixGRPO-Flash*. Notably, MixGRPO-Flash* enables acceleration over a larger number of ODE timesteps. A detailed description of the algorithm is provided in Appendix~\ref{app: mixGRPO-Flash_Algo}. 
These faster versions offer more substantial acceleration relative to base version of our \name{}, while simultaneously surpassing DanceGRPO in their alignment with human preferences.

\subsection{Coefficients-Preserving Sampling for Stable Exploration}
\label{sec:cps_method}
In addition to the mixed ODE-SDE design, we adopt Coefficients-Preserving Sampling (CPS)~\cite{wang2025coefficients} to improve sampling stability in the stochastic segment. Compared with standard SDE discretization, CPS better preserves the interpolation structure of flow matching while injecting controlled stochasticity, reducing high-frequency artifacts that can lead to reward hacking. In our implementation, CPS is used as an alternative stochastic discretization while keeping the same per-step Gaussian covariance scale $\sigma_t^2\Delta t\,\mathbf{I}$ for GRPO density-ratio/KL evaluation in Eq.~(9); therefore, the policy-ratio and KL parameterization remain unchanged. This yields cleaner rollouts and more reliable reward signals for GRPO optimization. Detailed derivations are provided in Appendix~\ref{app: CPS}.

\begin{table*}[!t]
  \centering
  \caption{Comparison between Mixed ODE-SDE and Global-SDE. \name{} achieves the best performance across multiple metrics. \name{}-Flash significantly reduces training time while outperforming DanceGRPO. 
  \textbf{Bold}: rank 1. \underline{Underline}: rank 2. $^*$The Frozen strategy means that optimization is only employed at the initial denoising steps.}
\vspace{-0.8em}
  \resizebox{0.9\linewidth}{!}{
    \begin{tabular}{lllccccc}
\toprule
\multirow{2}{*}{\textbf{Model}} & 
\multirow{2}{*}{\textbf{$\text{NFE}_{\pi_{\theta_{\text{old}}}}$}$\downarrow$} & \multirow{2}{*}{\textbf{$\text{NFE}_{\pi_{\theta}}$}$\downarrow$} &
\multirow{2}{*}{\textbf{Iteration Time~(s)}$\downarrow$}  & 
\multicolumn{4}{c}{\textbf{Human Preference Alignment}} \\
\cmidrule(lr){5-8}
& & & & \textbf{HPS-v2.1}$\uparrow$ & 
\textbf{Pick Score}$\uparrow$ & 
\textbf{ImageReward}$\uparrow$ & 
\textbf{Unified Reward}$\uparrow$ \\
\midrule

FLUX& / & / & / & 0.313 & 0.227 & 1.088 & 3.370 \\
\midrule
\multirow{3}{*}{DanceGRPO}
& 25 & 14  & 291.284   & 0.356 & 0.233 & 1.436 & 3.397 \\
& 25 & 4   & 149.978   & 0.334 & 0.225 & 1.335 & 3.374 \\
& 25 & 4$^*$ & 150.059 & 0.333 & 0.229 & 1.235 & 3.325 \\

\midrule
\rowcolor{green!10}
\name{} & 25 & 4 & 149.326 & \textbf{0.369} & \textbf{0.238} & \textbf{1.645} & \textbf{3.419} \\

\midrule
\rowcolor{green!10}
\name{}-Flash & 16 (Avg) & 4 & \underline{112.372}  & \underline{0.358} & \underline{0.236} & 1.528 & \underline{3.407} \\
\rowcolor{green!10}
\name{}-Flash* & 8 & 4$^*$ & \textbf{83.278} & 0.357 & 0.232 & \underline{1.624} & 3.402 \\
\bottomrule
\end{tabular}
  }
  \label{tab:main_multi_reward}
  \vspace{-1.0em}
\end{table*}
\section{Experiments}
\label{experiments}

\subsection{Experiment Setup}
\noindent\textbf{Dataset} \quad We conduct T2I experiments using the prompts provided by the HPDv2 dataset, which is the official dataset for the HPS-v2 benchmark~\cite{wu2023human}. The training set contains 103,700 prompts; in fact, \name{} already achieves strong human preference alignment before completing one full epoch, using only 9,600 prompts. The test set consists of 400 prompts. The prompts are diverse, encompassing four styles: ``Animation'', ``Concept Art'', ``Painting'', and ``Photo''.

\noindent\textbf{Base Model} \quad Following DanceGRPO~\cite{xue2025dancegrpo}, we use FLUX.1-dev~\cite{flux2024} as the primary backbone, a strong flow-matching-based text-to-image model. To further verify the robustness of \name{}, we also extend our method to additional backbones, including LoRA fine-tuning on SD3.5-M~\cite{esser2024scaling}. More implementation details are in Appendix~\ref{app: implementation_details}.

\noindent\textbf{Overhead Evaluation} \quad For the evaluation of overhead, we use two metrics: the number of function evaluations (NFE)~\cite{lu2022dpm} and the time consumption per iteration during training. The NFE is decomposed into $\text{NFE}_{\pi_{\theta_{\text{old}}}}$ and $\text{NFE}_{\pi_{\theta}}$. $\text{NFE}_{\pi_{\theta_{\text{old}}}}$ denotes the number of forward passes of the reference model used for rollout sampling. $\text{NFE}_{\pi_{\theta}}$ is the number of forward passes of the policy model solely for the policy ratio. Additionally, the average training time per GRPO iteration provides a more accurate reflection of the acceleration effect.

\noindent\textbf{Reward Model} \quad We employ four reward models: HPS-v2.1~\cite{wu2023human}, Pick Score~\cite{kirstain2023pick}, ImageReward~\cite{xu2023imagereward}, and Unified Reward~\cite{UnifiedReward}. Although all four are preference-aligned metrics, they capture complementary aspects of quality. In particular, ImageReward emphasizes image-text alignment and fidelity, while Unified Reward focuses more on semantic consistency. Following the findings of DanceGRPO~\cite{xue2025dancegrpo}, we adopt multi-reward guidance to obtain stronger and more stable alignment performance. To validate the generalization capability of \name{}, we conduct cross-domain evaluations across both reward models and datasets.

\subsection{Main Results}
\begin{figure}[t]
    \centering
    \includegraphics[width=0.88\linewidth]{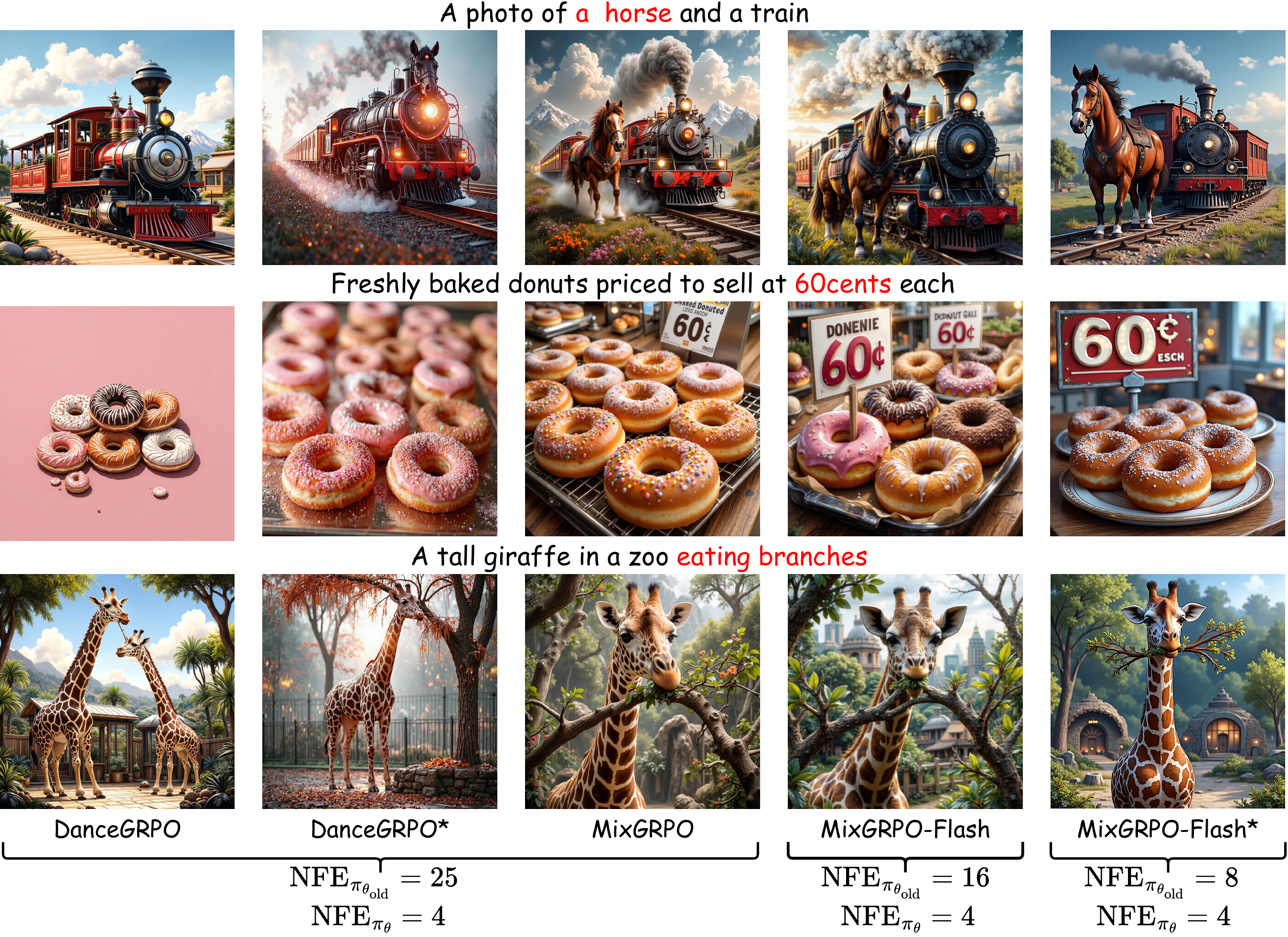}
    \vspace{-0.8em}
    
    \caption{Qualitative results of \name{}-Flash. As the training overhead is reduced, generated images maintain high quality.}
    \label{fig:multi_reward_performance_comparation_flash}
    \vspace{-0.8em}
\end{figure}

\subsubsection{Mixed ODE-SDE GRPO vs. Global-SDE GRPO} \quad We first evaluated the overhead and performance of \name{} against Global-SDE GRPO methods (e.g., DanceGRPO), with results shown in Table~\ref{tab:main_multi_reward}. In the official DanceGRPO setting, 14 timesteps are randomly selected from 25 for optimization. Our results show that \name{} achieves better performance while optimizing only 4 timesteps, reducing per-iteration GRPO training time from 291.284s to 150.839s. Qualitative results in Figure~\ref{fig:multi_reward_performance_comparation} further show that \name{} improves both aesthetic quality and semantic alignment. For fairness, we also modify DanceGRPO to use 4 timesteps (either randomly selected or the initial ones), matching the overhead of \name{}; however, DanceGRPO remains inferior to \name{} in alignment performance under both settings. For \name{}-Flash, we evaluate both progressive and frozen strategies. Although \name{}-Flash is slightly weaker than \name{}, all evaluation metrics are still higher than those of DanceGRPO (official setting), and the fastest setting reduces time from 291.284s to 83.278s. As shown in Figure~\ref{fig:multi_reward_performance_comparation_flash}, \name{}-Flash maintains strong image quality even as overhead changes. More visualized results are provided in Appendix~\ref{app: more_visualized_results}.

\subsubsection{Cross-domain Reward Models} \quad To evaluate the generalization ability of \name{}, we follow DanceGRPO and use HPS-v2.1 (aesthetics) and CLIP Score~\cite{hessel2021clipscore} (semantic-consistency) as in-domain reward models, under both single-reward and two-reward training settings. We further assess out-of-domain performance using Pick Score, ImageReward, and Unified Reward. As shown in Table~\ref{tab:main_single_or_two_reward}, \name{} consistently outperforms both DanceGRPO and the FLUX baseline on in-domain and out-of-domain metrics. These results indicate that \name{} improves overall image-generation alignment rather than relying on reward hacking (i.e., overfitting to in-domain reward models while failing to match human preference).

\subsubsection{\name{} vs. Other Alignment Methods} \quad In addition to DanceGRPO, we compare \name{} with other representative alignment methods on SD3.5-M, including DiffusionNFT~\cite{zheng2025diffusionnft}, Offline/Online Flow-DPO, and Flow-GRPO, with results shown in Table~\ref{tab:flow_grpo}. For fairness, the settings of the other alignment methods follow the best configurations reported in their original papers. We use HPS-v2.1, Pick Score, and ImageReward as multi-guidance and evaluation metrics. The results show that \name{} can converge to the best performance during training while optimizing only 4 timesteps. More experimental details are provided in Appendix~\ref{app: implementation_details}. We also compare with ReNO~\cite{eyring2024reno}, an alignment method for single-step generators; as shown in Appendix~\ref{app:reno_comp}, \name{} achieves better performance without incurring the overhead of step distillation.

\subsubsection{Extension to Larger-Scale Model and Video Generation} \quad We further validate the scalability of \name{} beyond standard T2I settings. On the industrial-scale HunyuanImage-3.0~\cite{cao2025hunyuanimage} backbone (80B parameters) trained on 512 GPUs, \name{} maintains stable optimization and substantial efficiency gains, demonstrating that its advantages persist at very large scale. Appendix~\ref{app:scaling_experiment} provides additional experimental details and visualizations of human blind-test preferences. We also extend \name{} to HunyuanVideo-1.5~\cite{hunyuanvideo2025} for text-to-video alignment, where it consistently outperforms Flow-GRPO in training stability and reward improvement. These results confirm that our mixed ODE-SDE design generalizes well across both larger image models and video generation tasks (see Appendix~\ref{app:t2v_extension} for training stability curves and more details).

\begin{table}[t]
    \centering
    \caption{Comparison under cross-rewards. The results demonstrate that \name{} achieves the best performance on both in-domain and out-of-domain rewards.}
    \vspace{-0.8em}
    \resizebox{0.9\linewidth}{!}{
        \begin{tabular}{lcccccc}
\toprule
\multirow{2}{*}{\textbf{Reward Model}} & 
\multirow{2}{*}{\textbf{Method}} & 
\multicolumn{2}{c}{\textbf{In Domain}} &
\multicolumn{3}{c}{\textbf{Out-of-Domain}} \\
\cmidrule(lr){3-4} \cmidrule(lr){5-7}
& & \textbf{HPS-v2.1} & \textbf{CLIP Score} & \textbf{Pick Score} & \textbf{ImageReward} & \textbf{Unified Reward} \\
\midrule
/ & FLUX & 0.313 & 0.388 & 0.227 & 1.088 & 3.370 \\
\midrule
\multirow{2}{*}{HPS-v2.1} 
& DanceGRPO & 0.367 & 0.349 & 0.227 & 1.141 & 3.270 \\
& \cellcolor{green!10}\name{} & \cellcolor{green!10}\textbf{0.373} & \cellcolor{green!10}\textbf{0.372} & \cellcolor{green!10}\textbf{0.228} & \cellcolor{green!10}\textbf{1.396} & \cellcolor{green!10}\textbf{3.370} \\
\midrule
\multirow{2}{*}{HPS-v2.1 \& CLIP Score} 
 & DanceGRPO & 0.346 & 0.400 & 0.228 & 1.314 & 3.377 \\
 & \cellcolor{green!10}\name{} & \cellcolor{green!10}\textbf{0.349} & \cellcolor{green!10}\textbf{0.415} & \cellcolor{green!10}\textbf{0.229} & \cellcolor{green!10}\textbf{1.416} & \cellcolor{green!10}\textbf{3.430} \\
\bottomrule
\end{tabular}
    }
    \vspace{-0.7em}
    \label{tab:main_single_or_two_reward}
\end{table}

\begin{table}[t]
    \centering
    \caption{Comparison with other alignment methods. The results demonstrate that \name{} outperforms others in training efficiency and performance.}
    \vspace{-0.74em}
    \resizebox{0.8\linewidth}{!}{
        \begin{tabular}{lccccc}
\toprule
\textbf{Model} & \textbf{$\text{NFE}_{\pi_{\theta_{\text{old}}}}$} & \textbf{$\text{NFE}_{\pi_{\theta}}$} & \textbf{HPS-v2.1} & \textbf{Pick Score} & \textbf{ImageReward} \\
\midrule
SD3.5-M & / & / & 0.307 & 0.227 & 1.163 \\
\midrule
Offline Flow-DPO & 40 & 40 & 0.304 & 0.222 & 1.452 \\ 
Online Flow-DPO & 40 & 40 & 0.313 & 0.221 & \textbf{1.500} \\
\midrule
DiffusionNFT & 10 & 10 & 0.313 & 0.235 & 1.494 \\
\midrule
Flow-GRPO & 10 & 10 & 0.331 & 0.232 & 1.457 \\
DanceGRPO & 10 & 10 & 0.309 & 0.226 & 1.433 \\
\rowcolor{green!10}\name{} & 10 & 4 & \textbf{0.342} & \textbf{0.236} & 1.485 \\
\bottomrule
\end{tabular}
    }
    \vspace{-1.4em}
    \label{tab:flow_grpo}
\end{table}

\subsection{Ablation Experiments}

\subsubsection{Sliding Window Hyperparamters} \quad As introduced in Section~\ref{sec: sliding_window_method}, the \textit{moving strategy}, \textit{shift interval} \( \tau \), \textit{window size} \( w \) and \textit{window stride} \( s \) are parameters introduced by the sliding window. We conducted ablation experiments on each of them.  For moving strategy, we compared three approaches: \textit{frozen}, where the window remains stationary; \textit{random}, where a random window position is selected at each iteration; and \textit{progressive}, where the sliding window moves incrementally with denoising steps. For \textit{progressive} strategy, we further evaluated different scheduling schemes in which the interval \( \tau \) starts from 25 and evolves over training. As shown in Table~\ref{tab:abl_SW_Strategy}, a constant schedule under the \textit{progressive} strategy yields the strongest overall trade-off. For \textit{shift interval} \( \tau \), we use \( \tau=25 \) as a robust default setting (see Table~\ref{tab:abl_SW_Iterations}). The number of forward passes for \( \pi_{\theta} \) increases with the growth of the window size \( w \), leading to greater time overhead. We compared different settings of \( w \), and results are shown in Table~\ref{tab:abl_SW_Size}. Ultimately, we use \( w = 4 \) as a balanced default setting between overhead and performance. For \textit{window stride} \( s \), we use \( s = 1 \) as the default choice, as shown in Table~\ref{tab:abl_SW_Step}.
\label{exp: abl_SW}

\begin{table*}[t]
    \centering
    \begin{minipage}[t]{0.49\textwidth}
        \centering
        \parbox[t][2.4em][t]{\linewidth}{\captionof{table}{Ablation for moving strategies.\label{tab:abl_SW_Strategy}}}
        \vspace{-0.6em}
        \resizebox{\columnwidth}{!}{
            \begin{tabular}{lcccccc}
\toprule
\textbf{Strategy} & \makecell{\textbf{Interval} \\ \textbf{Schedule}} & 
\textbf{HPS-v2.1} & \textbf{Pick Score} & \textbf{ImageReward} & \textbf{Unified Reward} \\
\midrule
Frozen      & /      & 0.354 & 0.234 & 1.580 & 3.403 \\
\midrule
Random      & Constant      & 0.365 & 0.237 & 1.513 & 3.388 \\
\midrule
\multirow{3}{*}{Progressive} & Decay (Linear) & 0.365 & 0.235 & 1.566 & 3.382 \\
            & Decay (Exp)    & 0.360 & \textbf{0.239} & \textbf{1.632} & 3.416 \\
            & Constant      & \textbf{0.367} & 0.237 & 1.629 & \textbf{3.418} \\
\bottomrule
\end{tabular}
        }
        \vspace{-0.3em}
    \end{minipage}\hfill
    \begin{minipage}[t]{0.49\textwidth}
        \centering
        \parbox[t][2.4em][t]{\linewidth}{\captionof{table}{Ablation for shift interval \( \tau \).\label{tab:abl_SW_Iterations}}}
        \vspace{-0.6em}
        \resizebox{0.8\columnwidth}{!}{
            \begin{tabular}{lcccc}
\toprule
\textbf{$\tau$} & \textbf{HPS-v2.1} & \textbf{Pick Score} & \textbf{ImageReward} & \textbf{Unified Reward} \\
\midrule
15 & 0.366 & 0.237 & 1.509 & 3.403 \\
20 & 0.366 & \textbf{0.238} & 1.610 & 3.411 \\
25 & \textbf{0.367} & 0.237 & \textbf{1.629} & \textbf{3.418} \\
30 & 0.350 & 0.229 & 1.589 & 3.385 \\
\bottomrule
\end{tabular}
        }
        \vspace{-0.3em}
    \end{minipage}
\end{table*}

\begin{table*}[t]
    \centering
    \begin{minipage}[t]{0.49\textwidth}
        \centering
        \captionof{table}{Ablation for window size $w$.}
        \vspace{-0.5em}
        \resizebox{0.85\columnwidth}{!}{
            \begin{tabular}{lccccc}
\toprule
\textbf{$w$} & $\text{NFE}_{\pi_{\theta}}$ & \textbf{HPS-v2.1} & \textbf{Pick Score} & \textbf{ImageReward} & \textbf{Unified Reward} \\
\midrule
1 & 1 & 0.359 & 0.234 & 1.629 & 3.235 \\
2 & 2 & 0.362 & 0.235 & 1.588 & \textbf{3.419} \\
4 & 4 & 0.367 & 0.237 & \textbf{1.629} & 3.418 \\
6 & 6 & \textbf{0.370} & \textbf{0.238} & 1.547 & 3.398 \\
\bottomrule
\end{tabular}
        }
        \vspace{-0.8em}
        \label{tab:abl_SW_Size}
    \end{minipage}\hfill
    \begin{minipage}[t]{0.49\textwidth}
        \centering
        \caption{Ablation for window stride \( s \).}
        \vspace{-0.5em}
        \resizebox{0.72\linewidth}{!}{
            \begin{tabular}{lcccc}
\toprule
\textbf{\( s \)} & \textbf{HPS-v2.1} & \textbf{Pick Score} & \textbf{ImageReward} & \textbf{Unified Reward} \\
\midrule
1 & 0.367 & 0.237 & \textbf{1.629} & \textbf{3.418} \\
2 & 0.357 & 0.236 & 1.575 & 3.391 \\
3 & \textbf{0.370} & 0.236 & 1.578 & 3.404 \\
4 & 0.368 & \textbf{0.238 }& 1.575 & 3.407 \\
\bottomrule
\end{tabular}
        }
        \vspace{-0.8em}
        \label{tab:abl_SW_Step}
    \end{minipage}
\end{table*}

\begin{table*}[t]
    \centering
    \setlength{\abovecaptionskip}{2pt}
    \setlength{\belowcaptionskip}{0pt}
    \begin{minipage}[t]{0.485\textwidth}
        \centering
        \captionof{table}{Comparison for different ODE solvers. The second-order Midpoint method achieves the best performance.}
        \resizebox{0.94\linewidth}{!}{
            \begin{tabular}{lccccc}
\toprule
\textbf{Order} & \textbf{Solver Type} & \textbf{HPS-v2.1} & \textbf{Pick Score} & \textbf{ImageReward} & \textbf{Unified Reward} \\
\midrule
1 & / & \textbf{0.367} & 0.236 & 1.570 & 3.403 \\
\midrule
\multirow{2}{*}{2} & Midpoint & 0.358 & \textbf{0.237} & \textbf{1.578} & \textbf{3.407} \\
  & Heun & 0.362 & 0.233 & 1.488 & 3.399 \\
\midrule
3 & / & 0.359 & 0.234 & 1.512 & 3.387 \\
\bottomrule
\end{tabular}
        }
        \label{tab:abl_Solver_order}
    \end{minipage}\hfill
    \begin{minipage}[t]{0.485\textwidth}
        \centering
        \captionof{table}{Comparison for sampling steps of the reference model. \name{}-Flash preserves performance while providing acceleration.}
        \resizebox{0.94\linewidth}{!}{
            \begin{tabular}{llccccc}
\toprule
\multirow{2}{*}{\textbf{Method}} &
\multicolumn{2}{c}{\textbf{Sampling Overhead}} &
\multicolumn{4}{c}{\textbf{Human Preference Alignment}} \\
\cmidrule(lr){2-3} \cmidrule(lr){4-7}
& \textbf{$\text{NFE}_{\pi_{\theta_\text{old}}}$} & \textbf{Time per Image (s)} & 
\textbf{HPS-v2.1} & \textbf{Pick Score} & \textbf{ImageReward} & \textbf{Unified Reward} \\
\midrule
DanceGRPO 
& 25 & 9.301 & 0.334 & 0.225 & 1.335 & 3.374 \\
\midrule
\multirow{3}{*}{MixGRPO-Flash} 
& 19 (Avg) & 7.343       & 0.357 & 0.236 & 1.564 & 3.394 \\
& 16 (Avg) & 6.426  & \textbf{0.362} & \textbf{0.237} & 1.578 & 3.407 \\
& 13 (Avg) & 5.453       & 0.344 & 0.229 & 1.447 & 3.363 \\
\midrule
\multirow{3}{*}{MixGRPO-Flash*} 
& 12 & 4.859 & 0.353 & 0.230 & 1.588 & 3.396 \\
& 10 & 4.214& 0.359 & 0.234 & 1.548 & \textbf{3.430} \\
& 8  & \textbf{3.789} & 0.357 & 0.232 & \textbf{1.624}  & 3.402 \\
\bottomrule
\end{tabular}
        }
        \vspace{-0.5em}
        \label{tab:abl_Solver_NFE}
    \end{minipage}
    \setlength{\abovecaptionskip}{10pt}
    \setlength{\belowcaptionskip}{0pt}
\end{table*}

\subsubsection{High Order ODE Solver} \quad MixGRPO-Flash, which incorporates a high-order ODE solver to accelerate the sampling process of the reference model, achieves an effective trade-off between speed and performance. For MixGRPO-Flash, we first conducted ablation experiments on the order of the solver, using DPM-Solver++~\cite{lu2022dpm++} as the high-order solver with the \textit{progressive} strategy. The results, as shown in Table~\ref{tab:abl_Solver_order}, indicate that the second-order mid-point method is the optimal setting, optimizing the most human preference alignment metrics while simultaneously accelerating the process.

Then we compared two acceleration approaches. One is \name{}-Flash, and the other is \name{}-Flash*. Both utilize a second-order ODE solver for acceleration, but they differ in their sliding-window moving strategies. The quantitative results are presented in Table~\ref{tab:abl_Solver_NFE}. \name{}-Flash requires the window to move throughout the training process, resulting in a smaller portion of the ODE being accelerated compared to \name{}-Flash*. Consequently, \name{}-Flash* not only achieves a higher degree of acceleration for the reference model but also yields superior results in the ImageReward~\cite{xu2023imagereward} and Unified Reward~\cite{UnifiedReward} metrics.

\subsubsection{Cross-Dataset Experiments and Sensitivity Analysis} 
\begin{figure}[h]
    \vspace{-0.8em}
    \centering
    \includegraphics[width=0.9\linewidth]{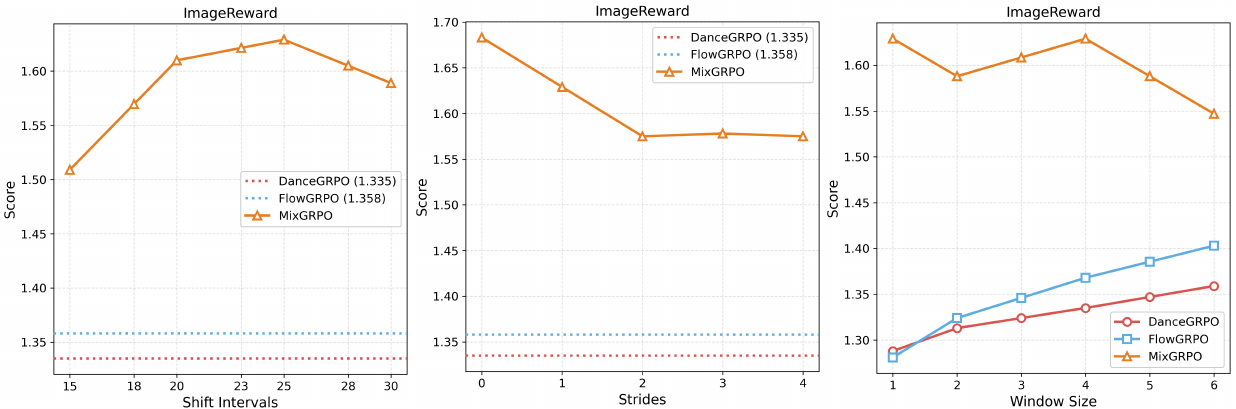}
    \caption{Sensitivity analysis of hyperparameters ($w$, $\tau$, $s$). The results show that \name{} is not sensitive to hyperparameters and consistently outperforms Global-SDE.}
    \label{fig:rebuttal_sensitivity}
\end{figure}

\quad 
Following the cross-dataset protocol in Appendix~\ref{app: cross-datset experiments}, we train on HPD-v2 and Pick-a-Pic-v1 datasets reciprocally and evaluate on both ID and OOD splits. The same hyperparameter setting (progressive-constant, $\tau=25$, $w=4$, $s=1$) remains consistently strong across datasets, while performance stays stable within a reasonable range of $(\tau, w, s)$ (Figure~\ref{fig:rebuttal_sensitivity}). These results indicate that \name{} is robust to dataset shifts and insensitive to delicate hyperparameter tuning, achieving reliable gains without additional tuning cost.

\subsubsection{Coefficients-Preserving Sampling} \quad
\label{exp:abl_cps}
We further ablate the stochastic sampler by comparing standard SDE sampling and CPS~\cite{wang2025coefficients} under the same training setup. As shown in Table~\ref{tab: sde_cps}, CPS consistently improves all reward metrics, indicating that reducing sampling artifacts provides more reliable optimization signals for preference alignment.

\begin{table}[t]
\centering
\caption{Comparison between SDE and CPS sampling, CPS effectively improves alignment performance.}
\vspace{-0.5em}
\resizebox{0.72\linewidth}{!}{
\begin{tabular}{lccccc}
\toprule
\textbf{Model} &  \textbf{HPS-v2.1} & \textbf{Pick Score} & \textbf{ImageReward} & \textbf{Unified Reward} \\
\midrule
FLUX & 0.313 & 0.227 & 1.088  & 3.370 &  \\
MixGRPO-SDE & 0.367   & 0.237 & 1.629 & 3.418  \\
MixGRPO-CPS  & \textbf{0.369}  & \textbf{0.238} & \textbf{1.645} & \textbf{3.419}\\
\bottomrule
\end{tabular}
}
\vspace{-0.5em}
\label{tab: sde_cps}
\end{table}
\normalsize
\section{Limitation}
\label{sec: limitation}
A key limitation of GRPO is that its performance ceiling is inherently tied to the capability of the reward model. Specifically, when the reward model is not sufficiently powerful to provide a comprehensive assessment of image quality, GRPO is prone to reward hacking, particularly in the later stages of training (see Appendix~\ref{app: mixsampling_for_hacking} for concrete examples). As confirmed by several studies~\cite{wu2025rewarddance, weng2024rewardhack, nishimura2024reward}, reward hacking is fundamentally driven by limitations of the reward model, rather than the RL algorithm itself. In this context, the primary goal of MixGRPO is not to eliminate these reward-model limitations, but to achieve faster convergence and stronger performance under the guidance of existing, imperfect reward models. In addition, our current sliding-window scheduler still relies on predefined hyperparameters ($w, \tau, s$). Although we observe strong robustness across datasets and backbones, an adaptive scheduler based on online signals (e.g., reward convergence or gradient variance) remains an important direction for future work. Looking ahead, while ultimate performance remains contingent on reward-model quality, we plan to develop and train stronger reward models in future work.

\section{Conclusion}
In this work, we have presented \name{}, a novel hybrid ODE-SDE framework for improving GRPO training efficiency and performance. We have proposed a strategy to confine optimization to a dynamic stochastic interval managed by a sliding window.
This guides the optimization process from broad exploration to fine-grained refinement, thus enhancing performance. Experiments demonstrate that \name{} has achieved superior performance in both single-reward and multi-reward settings while substantially reducing overhead. Furthermore, we have presented \name{}-Flash, a variant offering a flexible trade-off between performance and computational cost.
\label{page:main_end}

%
%
\bibliographystyle{splncs04}
\bibliography{main}

\clearpage
\setcounter{page}{1}
\begin{center}
\Large\textbf{Appendix of \name{}}
\end{center}

\section{Related Work}
\label{app: related_work}
\subsection{RL For Vision Generation}
Inspired by Proximal Policy Optimization (PPO)~\cite{schulman2017proximal}, early works~\cite{fan2023optimizing, black2023training, fan2023dpok, lee2023aligning} incorporated reinforcement learning (RL) into diffusion models by optimizing the score function~\cite{song2020score} via policy gradient methods, enabling image generation that better aligns with human preferences. Subsequently,~\cite{wallace2024diffusion} introduced offline Direct Preference Optimization (DPO) into text-to-image (T2I) generation, allowing diffusion models to learn directly from human preference pairs without explicit reward modeling, and demonstrating strong scalability on large models. In parallel, training-free methods~\cite{yeh2024training,tang2024inference,song2023loss,eyring2024reno} improve diffusion quality and efficiency by modifying inference-time sampling rather than updating model parameters. However, because offline-DPO may induce distributional shift by gradually moving the model away from its original data distribution through win--lose pair training, subsequent works~\cite{yuan2024self, liang2025aesthetic} adopt online alignment with step-aware preference signals to better preserve distributional consistency and improve optimization stability. More recently, GRPO-based approaches, \eg{} Flow-GRPO~\cite{liu2025flow}, and DanceGRPO~\cite{xue2025dancegrpo}, have further advanced RL-enhanced image generation. These methods extend Group Relative Policy Optimization (GRPO) to flow matching models by introducing stochasticity into the probability flow dynamics, enabling trajectory-level policy updates. This line of research has also inspired several variants, such as DiffusionNFT~\cite{zheng2025diffusionnft}, which reformulates diffusion reinforcement from a forward-process perspective. While effective, they rely on discretized sampling trajectories and full-step optimization, which introduce substantial computational overhead. Although DanceGRPO attempts to alleviate this issue by reducing or randomly subsampling optimization steps, the fundamental challenge of trajectory-level credit assignment remains. In this work, we revisit the probabilistic structure of diffusion probability flow and investigate the core mechanism of GRPO under mixed sampling and adaptive optimization schemes. 

\subsection{Sampling Methods for Probability Flow}
The development of generative sampling began with DDPM~\cite{ho2020denoising}, which used a slow, SDE-based probability flow requiring thousands of steps. To accelerate this, DDIM~\cite{song2020denoising} introduced a deterministic ODE-based approach, reducing sampling steps to around 100. This SDE / ODE duality was later unified by score-based models~\cite{song2020score}, paving the way for higher-order ODE solvers like DPM-Solver~\cite{lu2022dpm, lu2022dpm++} that reduced steps to 10. Higher-performance solvers~\cite{zheng2023dpm, zhao2023unipc} continue to be proposed; however, the gains are relatively marginal and have ultimately been replaced by the distillation method~\cite{salimans2022progressive, yin2024one}. Concurrently, flow matching models~\cite{lipman2022flow, esser2024scaling} simplified training by directly predicting the vector field, also enabling fast ODE sampling. Crucially, recent theoretical work~\cite{gao2025diffusionmeetsflow, albergo2023stochastic} has proven that flow matching and diffusion models share equivalent SDE and ODE formulations. This unification provides the theoretical foundation for our work, in which we explore an interleaved SDE and ODE sampling strategy within these probability flow models, thereby confining stochasticity to the SDE intervals in RL optimization, which shortens the effective MDP horizon and reduces training overhead.

\section{Proof of Convergence for Mixed ODE-SDE Sampling}
\label{app: proof_convergence}
To prove that the mixed ODE-SDE sampling method in Eq. (\ref{eq: mix_sampling}) has the same convergence as Eq. (\ref{eq: prob_ode}), which uses only ODE sampling, referencing \cite{song2020score}, we approach this from the perspective of distribution evolution, where the distribution at each time step, \eg $\frac{\partial{q_t(\mathbf{x})}}{\partial{t}}$ must be the same. Let the interval for SDE be defined as $S=\left[ t_l, t_r \right) \in \left[0,1\right)$. Along the denoising direction, when the same initial Gaussian noise distribution $q_0(\mathbf{x}_0)$ is given, the probability distribution evolution in the ODE interval preceding the SDE is completely identical. The key point is whether the distribution evolution of the SDE within the interval $S$ is completely equivalent to that of the ODE. If they are equivalent, then the ODE interval following the SDE will naturally be equivalent to using only ODE sampling. Next, we will provide a detailed proof for this key point.

Consider the SDE Eq. (\ref{eq: prob_sde}) in the interval $S$, which possesses the following form:
\begin{small}
\begin{equation}
\label{eq:app_sde_1}
    \mathrm{d}\mathbf{x} = [f(\mathbf{x}, t)-g^2(t)\nabla_{\mathbf{x}}\log{q_t(\mathbf{x})}]\mathrm{d}t + g(t)\mathrm{d}\mathbf{w}, \quad t \in S.
\end{equation}
\end{small}

The marginal probability density $q_t(\mathbf{x}_t)$ evolves according to Kolmogorov's equation (Fokker-Planck equation)~\cite{oksendal2003stochastic}
\begin{small}
\begin{equation}
\begin{aligned}
\label{eq:fp_eq}
    \frac{\partial q_t(\mathbf{x})}{\partial t} = & -\nabla_\bfx \cdot \big[\big(f(\bfx, t)-g^2(t)\nabla_\bfx\log q_t(\bfx)\big) q_t(\bfx)\big]  \\ &+ \frac{1}{2}g^2(t)\nabla^2_\bfx q_t(\bfx)
\end{aligned}
\end{equation}
\end{small}

According to the definition of the Laplace operator$\nabla^2h \equiv \nabla \cdot \nabla(h)$ and $\nabla_\bfx \log q_t(\bfx) = \frac{\nabla q_t(\bfx)}{q_t(\bfx)}$, we can obtain:
\begin{small}
\begin{equation}
\label{eq:fp_eq_1}
\begin{aligned}
    \frac{\partial q_t(\mathbf{x})}{\partial t} 
    &= -\nabla_\bfx \cdot \big[f(\bfx, t) q_t(\bfx) -g^2(t)\nabla_\bfx q_t(\bfx)\big] \\ &\quad~ + \frac{1}{2}g^2(t)\nabla^2_\bfx q_t(\bfx)  \\
    &= -\nabla_\bfx \cdot [f(\bfx,t)q_t(\bfx) - \frac{1}{2}g^2(t)\nabla_\bfx q_t(\bfx)] \\
    &= -\nabla_\bfx \! \cdot \! \big[ \! \underbrace{ \big( f(\bfx,t) \! - \! \frac{1}{2}g^2(t)\nabla_\bfx \log q_t(\bfx) \big) }_{f_{\text{ODE}}(\bfx,t)} q_t(\bfx) \! \big].
\end{aligned}
\end{equation}
\end{small}
The Eq. (\ref{eq:fp_eq_1}) is indeed the Fokker-Planck equation of the ODE Eq. (\ref{eq: prob_ode}). Therefore, within the interval \( S \), the distribution evolution of SDE and ODE sampling is consistent.

\section{DPM-Solver++ for Recitified Flow}
\label{app: dpm_for_rf}

For clarity and to avoid ambiguity between continuous time and discrete steps, we adopt the following notation in this section. We denote the discrete time steps by an index $i \in \{0, 1, \dots, T-1\}$, where $T$ is the total number of sampling steps. The continuous time corresponding to step $i$ is denoted by $t_i = \frac{i}{T} \in [0, 1)$.

The DPM-Solver++ algorithm~\cite{lu2022dpm++} is originally designed for the $\mathbf{x}_0\text{-}prediction$ diffusion model~\cite{rombach2022high}, where the model outputs the denoised feature \( \mathbf{x}_0 \) based on the noisy feature \( \mathbf{x}_{t_i} \), the time condition \( t_i \) and the text condition $c$. 
According to the definition of Rectified Flow (RF)~\cite{liu2022flow}, there is the following transfer equation:
\begin{equation}
\label{eq: rf_transfer}
    \mathbf{x}_{t_i} = t_i \mathbf{x}_1 + (1-t_i)\mathbf{x}_0.
\end{equation}

According to the theory of stochastic interpolation~\cite{albergo2023stochastic}, RF effectively approximates \( \mathbf{x}_1 - \mathbf{x}_0 \) by modeling \( \mathbf{v}_{t_i} \):
\begin{equation}
\label{eq: stocha_inter}
    \mathbf{v}_{t_i}=\mathbf{x}_1-\mathbf{x}_0.
\end{equation}

Based on Eq. (\ref{eq: rf_transfer}) and Eq. (\ref{eq: stocha_inter}), we obtain the following relationship:
\begin{equation}
    \mathbf{x}_0=\mathbf{x}_{t_i}-\mathbf{v}_{t_i} t_i.
\end{equation}
By using a neural network for approximation, we establish the relationship between RF and the $\mathbf{x}_0\text{-}prediction$ model:
\begin{equation}
    \mathbf{x}_\theta(\mathbf{x}_i, t_i, c) = \mathbf{x}_i - \mathbf{v}_\theta(\mathbf{x}_i, t_i, c) \cdot t_i.
\end{equation}

Taking the multistep second-order DPMSolver++ as an example (see Algorithm 2 in ~\cite{lu2022dpm++}), we derive the corrected \(\mathbf{x}_\theta\) for the RF sampling process as \(\mathbf{D}_i\):
\begin{equation}
\label{eq: dpm_difference_for_rf}
\begin{aligned}
    \mathbf{D}_i 
    \gets 
    \left(1+\frac{h_i}{2h_{i-1}}\right) \left( \mathbf{x}_{i-1} - \mathbf{v}_\theta(\mathbf{x}_{i-1}, t_{i-1}, c) \cdot t_{i-1} \right) 
    \\  
    \quad - \frac{h_i}{2h_{i-1}} \left( \mathbf{x}_{i-2} - \mathbf{v}_\theta(\mathbf{x}_{i-2}, t_{i-2}, c) \cdot t_{i-2} \right),
\end{aligned}
\end{equation}
where $h_i=\lambda_{t_i} - \lambda_{t_{i-1}}$. The continuous time $t_i$ corresponds to the discrete step $i$ over a total of $T$
sampling steps. The term $\lambda_{t_i}$ is the log-\textit{signal-to-noise-ratio} (log-SNR) and is defined in RF as:
\begin{equation}
    \lambda_{t_i} := \log \left( \frac{1-t_{i}}{t_i} \right).
\end{equation}

Based on the exact discretization formula for the \textit{probability flow ODE} proposed in DPM-Solver++ (Eq. (9) in \cite{lu2022dpm++}), we can derive the final transfer equation:
\begin{small}
\begin{equation}
\label{eq: dpm_for_rl_transfer}
    \mathbf{x}_i 
    \gets 
    \frac{t_i}{t_{i-1}} \mathbf{x}_{i-1}
    - (1-t_i)\left( e^{-h_i} - 1 \right) \mathbf{D}_i, \quad 1\leq i<T.
\end{equation}
\end{small}

\section{MixGRPO-Flash Algorithm}
\label{app: mixGRPO-Flash_Algo}
\begin{algorithm}[!t]
\caption{\name{}-Flash Training Process}
\label{algo:mixgrpo-flash}
\small
\begin{algorithmic}[1]
\Require initial policy model $\pi_\theta$; reward models $\{R_k\}_{k=1}^K$; prompt dataset $\mathcal{C}$; total sampling steps $\tilde{T}$; number of samples per prompt $N$; ODE compression rate $\tilde{r}$
\Require sliding window $W(l)$, window size $w$, shift interval $\tau$, window stride $s$
\State Init left boundary of $W(l)$: $l \gets 0$
\For{training iteration $m=1$ \textbf{to} $M$}
    \State Sample batch prompts $\mathcal{C}_b \sim \mathcal{C}$
    \State Update old policy model: $\pi_{\theta_{\text{old}}} \gets \pi_\theta$
    \For{each prompt $\mathbf{c} \in \mathcal{C}_b$}
        \State Init the same noise $\mathbf{x_0} \sim \mathcal{N}(0,\mathbf{I})$
        \For{generate $i$-th image from $i=1$ \textbf{to} N}
            \For{sampling timestep $t=0$ \textbf{to} $\tilde{T}-1$}
                \If{$t<l$}
                    \State Use first-order ODE sampling to get $\mathbf{x}^{i}_{t+1}$
                \ElsIf{~$l\leq t < l+w$}
                    \State Use SDE sampling to get $\mathbf{x}^{i}_{t+1}$ 
                \Else \Comment{higher-order ODE}
                    \State Use DPM-Solver++ sampling to get $\mathbf{x}^{i}_{t+1}$
                \EndIf
            \EndFor
        \EndFor
        \For{$i$-th image from $i=1$ \textbf{to} N}
            \State Calculate advantage: 
            $A_i \gets \sum_{k=1}^K \frac{R(\mathbf{x}^i_{\tilde{T}}, \mathbf{c})^i_k - \mu_k}{\sigma_k}$ 
        \EndFor
        \For{optimization timestep $t \in W(l)$}
            \State Update policy model: $\theta \gets \theta + \eta\nabla_\theta\mathcal{J}$
        \EndFor
    \EndFor
    \If{use \name{}-Flash*} \Comment{move sliding window}
        \State $l \gets 0$
    \Else
        \If{$m \bmod \tau~\text{is}~0$} 
            \State $l \gets \min(l+s,~T-w)$ 
        \EndIf
    \EndIf
\EndFor
\end{algorithmic}

\end{algorithm}

\name{}-Flash Algorithm~\ref{algo:mixgrpo-flash} accelerates the ODE sampling that does not contribute to the calculation of the policy ratio after the sliding window by using DPM-Solver++ in the Eq. (\ref{eq: dpm_for_rl_transfer}). We introduce a compression rate \( \tilde{r}\) such that the ODE sampling after the window only requires \( (T - l - w) \tilde{r} \) time steps. And the total time-steps is $\tilde{T}= l+w+ (T - l - w) \tilde{r}$ The final algorithm is as follows:

Note that when using \name{}-Flash*, the frozen strategy is applied, with the left boundary of the sliding window \( l \equiv 0 \). The theoretical speedup of the training-time sampling can be described as follows:
\begin{equation}
    S=\frac{T}{w+(T-w)\tilde{r}}.
\end{equation}

For \name{}-Flash, since the sliding window moves according to the progressive strategy during training, the average speedup can be expressed in the following form:
\begin{small}
\begin{equation}
    S=\frac{T}{\mathbb{E}_l \left( w+l+ \lceil (T-w-l) \tilde{r} \rceil \right)} < \frac{T}{w+(T-w)\tilde{r}}.
\end{equation}
\end{small}

\section{Implementation Details}
\label{app: implementation_details}

\textbf{Based on FLUX.1-dev~\cite{flux2024}} Training is conducted on 32 NVIDIA GPUs with a per-GPU batch size of 1 for up to 300 iterations. For training-time sampling, we first apply the time shift $\tilde{t}=\frac{t}{1-(\tilde{s}-1)t}$ to uniformly sampled timesteps $t_i=\frac{i}{T}$ ($i\in[0,\ldots,T-1]$), and define $\sigma_t=\eta\sqrt{\frac{\tilde{t}}{1-\tilde{t}}}$ with $\tilde{s}=3$ and $\eta=0.7$. We set $T=25$ sampling steps. In GRPO training, the model generates 12 images per prompt, clips advantages to $[-5,5]$, and uses 3-step gradient accumulation (i.e., 4 gradient updates per training iteration). For multi-reward training, all rewards are equally weighted. We use AdamW~\cite{loshchilov2017decoupled} with a learning rate of $1\times10^{-5}$ and weight decay of $1\times10^{-4}$, train in bf16 mixed precision, and keep master weights in fp32.

\noindent\textbf{Based on SD3.5-M~\cite{esser2024scaling}} Training is performed on 24 NVIDIA H20 GPUs using Adam with $\mathrm{lr}=3\times10^{-4}$ and $\mathrm{wd}=1\times10^{-4}$. We sample $N=24$ images per prompt at $512\times512$ resolution, use $T=10$ steps for training and $T=40$ for evaluation, and set the classifier-free guidance scale to 4.5. The global batch size is 96 prompts per update (12 prompts per GPU with 6 gradient accumulation steps). We use a timestep fraction of 0.99, KL coefficient $\beta=0.001$, and EMA. For MixGRPO, we set window size $w=4$, shift interval $\tau=150$, and stride $s=1$. Rewards are an equally weighted combination of HPSv2, PickScore, and ImageReward to jointly optimize aesthetics and text-image alignment.

\section{Multi-step MDP GRPO vs. Alignment with Single-step Generators}
\label{app:reno_comp}
To evaluate the efficiency of \name{} under multi-step MDP rollout against models that leverage single-step generators, we include a comparison with Reward-based Noise Optimization (ReNO)~\cite{eyring2024reno}. ReNO improves one-step text-to-image models by iteratively optimizing the initial latent noise during inference with reward gradients from frozen preference models.

Nevertheless, the application of ReNO is largely restricted to single-step models (e.g., SD3.5-Large-Turbo~\cite{esser2024scaling}) due to the cost of backpropagating through the full sampling path. In contrast, our \name{} framework is compatible with standard multi-step diffusion and flow-matching pipelines. As shown in Table~\ref{tab: reno_comp}, \name{} provides a better efficiency--alignment trade-off while avoiding the additional step-distillation requirement.

\begin{table}[t]
\centering
\caption{Comparison with ReNO~\cite{eyring2024reno}. \name{} enables efficient optimization of the base model, whereas ReNO requires step distillation.}
\resizebox{0.8\linewidth}{!}{
\begin{tabular}{lccccc}
    \toprule
    \textbf{Base Model} &
    \textbf{Method} & 
    \textbf{Training Overhead (sec/step)$\downarrow$} &
    \textbf{HPS-v2.1$\uparrow$} & 
    \textbf{Pick Score$\uparrow$} & \textbf{ImageReward$\uparrow$} \\
    \midrule
    \multirow{2}{*}{SD3.5-L} & ReNO & $\gg 94.297^\dagger$ & 0.352 & 0.235 & \textbf{1.725} \\
    & MixGRPO & \textbf{94.297} & \textbf{0.366}  & \textbf{0.237}  & 1.659 \\
    \bottomrule
    \multicolumn{6}{l}{\footnotesize \textbf{Bold}: best results per base model; $^\dagger$ReNO requires one-time distillation overhead to a single-step generator.}
\end{tabular}
}
\label{tab: reno_comp}
\end{table}

\section{Degradation caused by accelerating the ODE before the SDE window}
\label{app: before_ODE_error}

\begin{figure}[t]
    \centering
    \includegraphics[width=\linewidth]{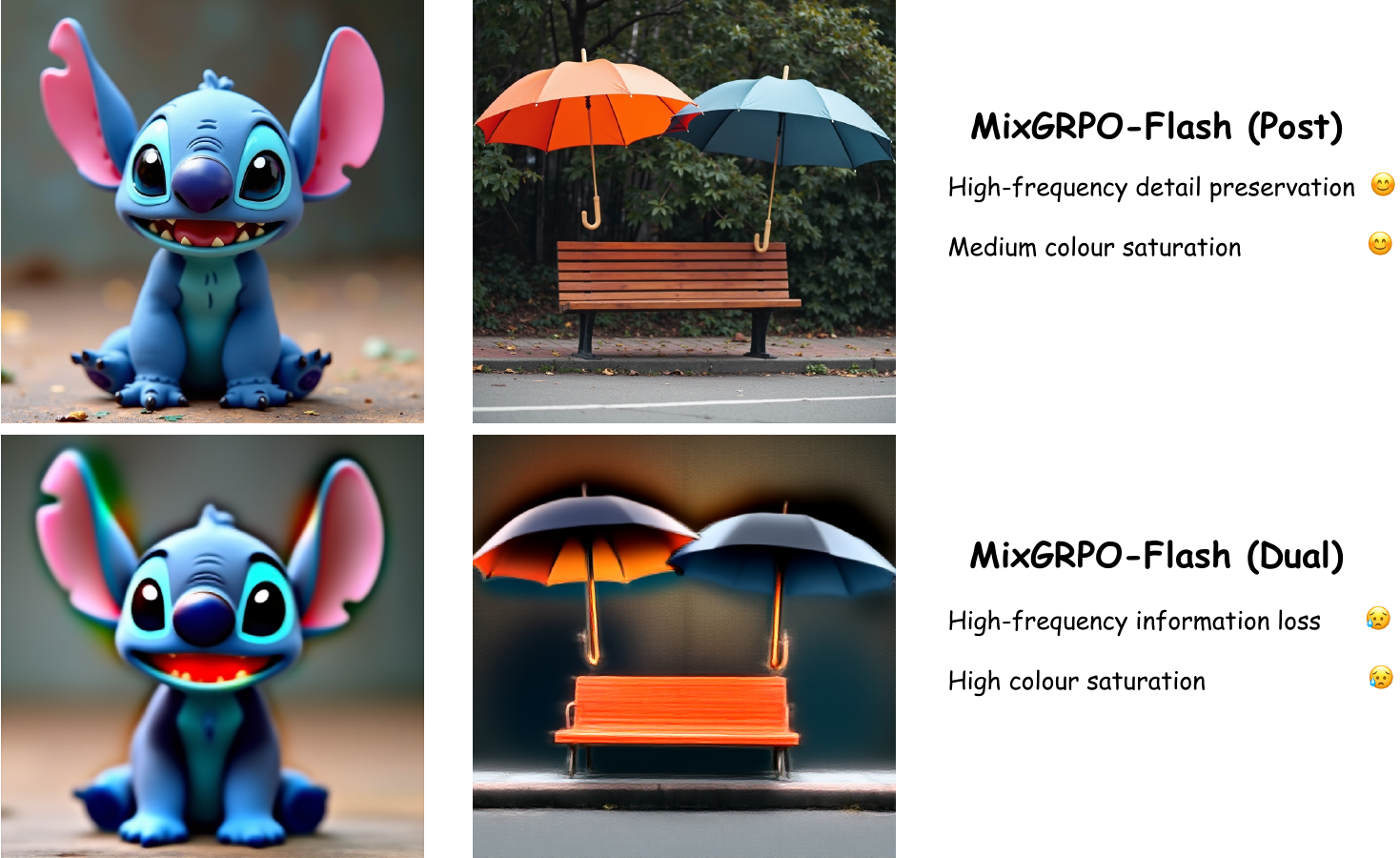}
    \caption{GRPO rollout visualizations of MixGRPO-Flash (Dual) and MixGRPO-Flash (Post).}
    \label{fig:dual_vs_post_visual}
\end{figure}

In the main paper, we accelerate ODE segments outside the optimization window with \name{}-Flash to reduce training-time sampling cost. A key implementation choice is \textit{where} to apply this acceleration relative to the SDE optimization window: \textbf{Dual} (accelerating both the pre-window and post-window ODE segments) versus \textbf{Post} (accelerating only the post-window ODE segment). We observe that accelerating the ODE segment \textit{before} the SDE window causes visible degradation. As shown in Fig.~\ref{fig:dual_vs_post_visual}, MixGRPO-Flash (Post) shows better high-frequency detail preservation and medium colour saturation, while MixGRPO-Flash (Dual) exhibits high-frequency information loss and high colour saturation. Moreover, this degradation pattern of MixGRPO-Flash (Dual) becomes increasingly severe as training proceeds.

For a controlled comparison, all settings are identical between the two variants except the acceleration strategy (i.e., whether pre-window ODE steps are accelerated): the same base model initialization, prompts, reward models, optimizer and hyperparameters, rollout length, sliding-window configuration, and training iterations. The visual examples in Fig.~\ref{fig:dual_vs_post_visual} are GRPO rollout sampling results. Quantitative results in Table~\ref{tab:dual_vs_post} further support the visual findings, where MixGRPO-Flash (Post) consistently outperforms MixGRPO-Flash (Dual) on HPS-v2.1, Pick Score, and ImageReward. This indicates that keeping the pre-window ODE segment uncompressed is important for stable GRPO optimization, because it determines the state distribution entering the RL-optimized SDE window.

\begin{table}[t]
\centering
\caption{Comparison of MixGRPO-Flash variants with and without pre-window ODE acceleration. Accelerating the dual ODE segment causes GRPO optimization collapse.}
\resizebox{0.8\linewidth}{!}{
\begin{tabular}{lccccc}
    \toprule
    \textbf{Base Model} &
    \textbf{Method} &
    \textbf{HPS-v2.1$\uparrow$} & 
    \textbf{Pick Score$\uparrow$} & \textbf{ImageReward$\uparrow$} \\
    \midrule
    \multirow{2}{*}{FLUX} & MixGRPO-Flash~(Dual)  & 0.335 & 0.223 & 1.235 \\
    & MixGRPO-Flash~(Post) & \textbf{0.358}  & \textbf{0.236}  & \textbf{1.528} \\
    \bottomrule
\end{tabular}
}
\label{tab:dual_vs_post}
\end{table}

\section{Scaling MixGRPO to Industrial-Scale Models}
\label{app:scaling_experiment}

To further validate the scalability and practical impact of \name{}, we extended our experiments to an industrial-scale setting, moving beyond standard academic benchmarks. \textcolor{black}{This large-scale evaluation aims to demonstrate that \name{} is not merely a refinement of existing techniques but a fundamental paradigm shift in efficient reinforcement learning for flow-based generative models.}

\textbf{Experimental Setup.} We applied \name{} to \textbf{HunyuanImage-3.0}~\cite{cao2025hunyuanimage}, an advanced flow-based large-scale text-to-image model with approximately \textbf{80B parameters}. The training was conducted on a high-performance computing cluster using \textbf{512 NVIDIA GPUs}. This unprecedented scale allows us to examine the behavior of our Mixed ODE-SDE paradigm and sliding window strategy when dealing with extremely high-dimensional parameter spaces and massive data throughput.

\begin{figure}[h]
    \centering
    \includegraphics[width=\linewidth]{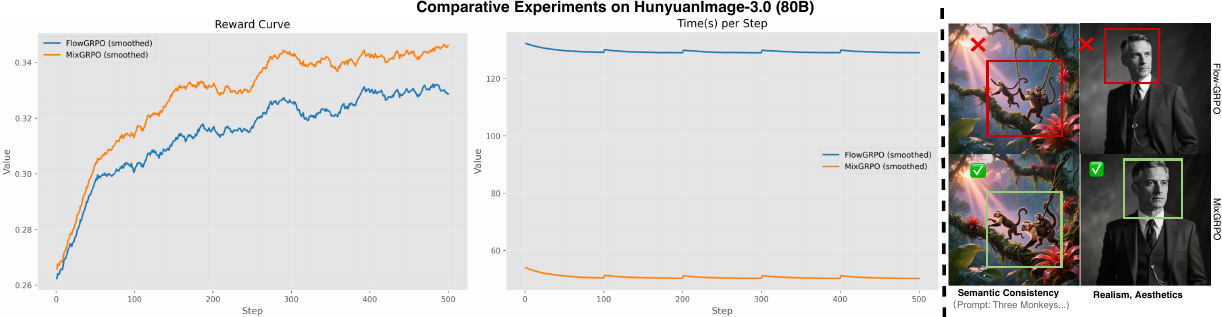}
    \caption{Performance and efficiency gains of MixGRPO when scaled to the 80B HunyuanImage-3.0 model on 512 GPUs. The results demonstrate that our method maintains its significant advantages in both convergence speed and reward optimization even at an industrial scale.}
    \label{fig:scaling_results}
\end{figure}

\textbf{Efficiency and Performance Gains.} As illustrated in Fig.~\ref{fig:scaling_results}, \name{} achieves remarkable performance improvements while significantly reducing the computational burden. Specifically, our method provides a consistent speedup (approximately 70\%) compared to full-trajectory optimization methods like FlowGRPO or DanceGRPO. More importantly, the convergence of \name{} at this scale is notably more stable. By operationalizing RL Temporal Discounting within the flow-matching framework, our sliding window strategy effectively resolves the gradient conflicts that often plague full-trajectory RL in extremely large models.

\textbf{Methodological Impact.} The success of these industrial-scale experiments underscores several key advantages of \name{}:
\begin{itemize}
    \item \textbf{Optimal Budget Allocation:} \name{} offers a transformative perspective on how to allocate the sampling and optimization budget for continuous-time generation. Instead of treating all timesteps equally, it focuses the RL signal where it is most effective.
    \item \textbf{Viability of GRPO:} These results transform GRPO from a theoretically interesting but computationally prohibitive technique into a viable, cost-effective paradigm for the next generation of 100B+ parameter models.
    \item \textbf{Robustness to Scale:} The performance gains do not diminish as the model size or GPU count increases, proving that the theoretical foundation of our Mixed ODE-SDE paradigm (detailed in Sec.~\ref{sec: mixed ODE-SDE}) is rigorously sound and scalable.
\end{itemize}

In summary, the application of \name{} to the 80B HunyuanImage-3.0 model confirms that our approach represents a fundamental breakthrough in the efficiency and effectiveness of Reinforcement Learning from Human Feedback (RLHF) for state-of-the-art generative models.

\section{Extension to Text-to-Video Generation}
\label{app:t2v_extension}

\begin{figure*}[t!]
    \centering
    \includegraphics[width=\linewidth]{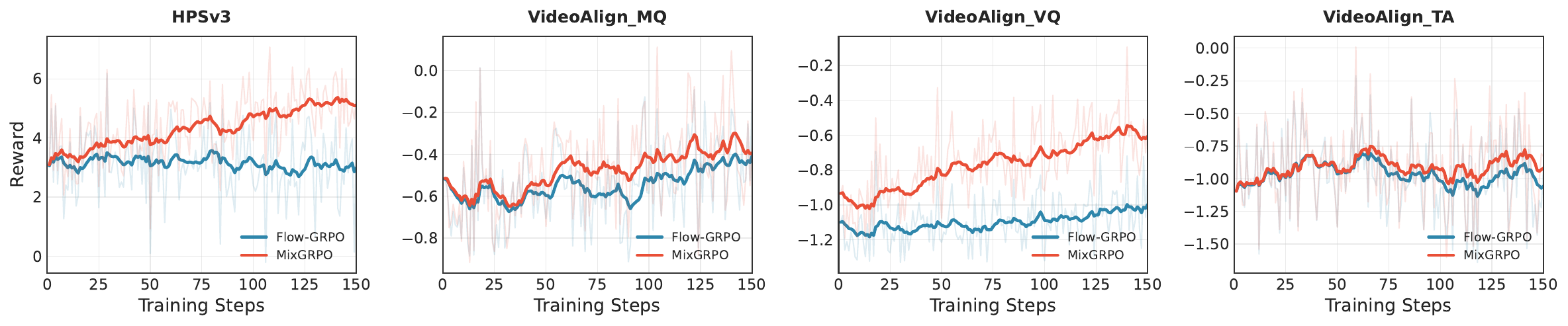}
    \caption{\textbf{Training Reward Dynamics on HunyuanVideo-1.5.} Comparison of reward curves between MixGRPO (ours) and Flow-GRPO during text-to-video alignment training. }
    \label{fig:video_reward}
\end{figure*}

To further validate the scalability and robustness of MixGRPO, we conducted comparative experiments against Flow-GRPO using the HunyuanVideo-1.5~\cite{hunyuanvideo2025} backbone. 

\paragraph{Implementation Details.}
For the experimental setup, training is performed on 64 Nvidia H800 GPUs. Consistent with the pre-training strategy, we utilize the Muon~\cite{jordan2024muon} optimizer ($\text{lr}=1\text{e-}5$, $\text{wd}=0.01$). During the rollout phase, we adopt prompts from DanceGRPO~\cite{xue2025dancegrpo}, generating $N=8$ video samples per prompt at a resolution of $480 \times 864$ with 121 frames (approx. 5 seconds). The sampling process is configured with $T=25$ steps, a time shift of 5, and a stochastic scale $\eta=0.5$. The global batch size is set to 8 prompts (resulting in a total of 64 video samples) for each update. Regarding MixGRPO-specific configurations, we set the sliding window size $w=6$, shift interval $\tau=100$, and stride $s=1$. The optimization is guided by an equal-weighted combination of HPSv3~\cite{ma2025hpsv3widespectrumhumanpreference} and VideoAlign~\cite{liu2025improving}, comprehensively assessing aesthetics, semantics, and motion quality.

\paragraph{Analysis of Training Dynamics.}
The training reward curves are shown in Figure~\ref{fig:video_reward}. We observe that standard Flow-GRPO is unstable and inefficient in the high-dimensional latent space of video generation. Specifically, on VideoAlign sub-metrics (Visual Quality, VQ; Text Alignment, TA) and HPSv3, Flow-GRPO yields only marginal gains and even degrades at some stages. 
In contrast, MixGRPO demonstrates superior stability and convergence efficiency. By confining the stochastic exploration to a sliding window, MixGRPO effectively reduces the optimization search space, allowing for more precise gradient updates. As shown in the figures, MixGRPO achieves consistent monotonic improvements across all three metrics—Aesthetics (HPSv3), Motion Quality (MQ) and Visual Quality (VQ)—significantly outperforming the baseline. This confirms that our mixed ODE-SDE strategy is not only effective for images but crucial for stabilizing RL training in complex video generation tasks.

\section{Cross-dataset Experiments and Hyperparameter Sensitivity Analysis}
\label{app: cross-datset experiments}

To investigate the robustness and parameter sensitivity of the sliding window strategy in \name{}, we conducted a series of cross-dataset ablation studies. We established two reciprocal settings to evaluate both in-domain (ID) and out-of-domain (OOD) performance. In cross-dataset experiment 1, the model was trained on the HPDv2~\cite{wu2023human} dataset and evaluated on the test sets of HPDv2 (ID) and Pick-a-Pic v1 (OOD). In cross-dataset experiment 1, conversely, the model was trained on Pick-a-Pic v1~\cite{kirstain2023pick} and evaluated on the test sets of Pick-a-Pic v1 (ID) and HPDv2 (OOD). Within these settings, we ablated the key parameters of MixGRPO: \textbf{the moving strategy, shift interval $\tau$, window size $w$ and window stride $s$}. The results are presented in the Tables~\ref{tab:ood_strategy},~\ref{tab:ood_iterations},~\ref{tab:ood_size},~\ref{tab:ood_step}.

As shown in Table~\ref{tab:ood_strategy}, Progressive-Decay(Exp) and Progressive-Constant emerge as the top-performing moving strategies. While Progressive-Decay(Exp) consistently achieves the highest ImageReward~\cite{xu2023imagereward} score in in-domain evaluations, Progressive-Constant demonstrates superior overall performance. Notably, the Progressive-Constant strategy consistently outperforms all others in out-of-domain tests, demonstrating its strong robustness and generalization capabilities.

The results concerning the shift interval ($\tau$), which dictates the frequency of window movement in training iteration steps, are summarized in Table~\ref{tab:ood_iterations}. Overall, the optimal performance is consistently observed at $\tau=25$. Specifically, as $\tau$ is incrementally increased from $15$ to $25$, performance metrics exhibit a gradual rise across both in-domain and out-of-domain test sets. However, a crucial observation is the sharp decline in performance when the interval is extended to $\tau=30$, with results even falling below the baseline achieved at $\tau=15$. This trend suggests that a moderate reduction in the sliding window's movement speed (from $\tau=15$ to $\tau=25$) allows the temporal behavior within the window to be sufficiently optimized. Conversely, overly slow window movement (exceeding $\tau=25$) is prone to over-optimization at certain timesteps, causing the model's distribution to diverge significantly from the target preferences of the reward models. The cross-dataset experiments further confirm the universality of $\tau=25$ as the optimal setting.

The comparative results for the window size ($w$) are presented in Table~\ref{tab:ood_size}. It is evident that both $w=4$ and $w=6$ represent optimal settings. Specifically, $w=6$ slightly surpasses $w=4$ only on the in-domain test set of Cross-dataset Experiment 1, and in other experiments, it only holds a marginal advantage in the HPS-v2.1~\cite{wu2023human} score. It is crucial to note that the window size $w$ corresponds to the number of denoising timesteps that must be optimized during each training iteration, which, in turn, is linearly correlated with the overall optimization overhead of the Reinforcement Learning (RL) process. Consequently, $w=4$ is identified as the best compromise, offering an optimal trade-off between performance and computational efficiency. Regarding parameter sensitivity, we observe that the change in performance metrics—across both in-domain and out-of-domain tests—is relatively minimal when $w$ varies between $2$ and $6$. This stability suggests that the model exhibits robust performance with respect to the window size setting. As long as the window size remains within a reasonable range (e.g., $2 \le w \le 6$), the model is capable of effective learning and maintaining stable performance, thereby reducing necessity for meticulous hyperparameter tuning.

The results concerning the stride ($s$), which defines the step size for each sliding window movement, are presented in Table~\ref{tab:ood_step}. Considering all experimental metrics across the different datasets, a stride of $s=1$ is identified as the optimal setting overall. It is important to note that the optimal window size of $w=4$ was used in this experiment. With $s=1$, the last three timesteps within the window are repeatedly optimized in the subsequent training cycle. Crucially, the experimental results indicate that this repetition leads to performance improvement. This is likely due to the nature of the high-SNR denoising process, where the initial timesteps (e.g., the first timestep in the window) are subject to a greater extent of GRPO optimization, despite the window size being set at $w=4$. Conversely, the behavior learned during the subsequent three timesteps remains relatively under-optimized (underfitted) and thus significantly benefits from this repeated optimization. Furthermore, we observe that the highest in-domain HPS-v2.1~\cite{wu2023human} score is achieved when $s=3$. This localized optimal result may be attributed to the HPS-v2.1~\cite{wu2023human} reward model's preference alignment being relatively easier to optimize, suggesting that repeated optimization (i.e., smaller strides) could potentially induce over-optimization in this specific metric.In summary, $s=1$ proves to be the most robust and generalizable choice across the varying experimental conditions.

\textbf{Shift Interval ($\tau$).} The results concerning the shift interval ($\tau$), which dictates the frequency of window movement in training iteration steps, are summarized in Table~\ref{tab:ood_iterations}. Overall, the optimal performance is consistently observed at $\tau=25$. Specifically, as $\tau$ is incrementally increased from $15$ to $25$, performance metrics exhibit a gradual rise across both ID and OOD test sets. However, a crucial observation is the sharp decline in performance when the interval is extended to $\tau=30$, with results even falling below the baseline achieved at $\tau=15$. This trend suggests that a moderate reduction in the sliding window's movement speed allows the temporal behavior within the window to be sufficiently optimized. Conversely, overly slow window movement (exceeding $\tau=25$) is prone to \textbf{reward overfitting} at certain timesteps, causing the model's distribution to diverge significantly from the target preferences of the reward models. The cross-dataset experiments further confirm the universality of $\tau=25$ as the optimal setting across different training sources.

\textbf{Window Size ($w$) and Stride ($s$).} The comparative results for window size ($w$) and stride ($s$) are presented in Tables~\ref{tab:ood_size} and \ref{tab:ood_step}. As visualized in Fig.~\ref{fig:rebuttal_sensitivity}, we observe three critical trends: 
(1) \textbf{Consistent Superiority:} MixGRPO's performance (solid lines) consistently stays above the DanceGRPO and FlowGRPO baselines (dashed lines) regardless of the specific values of $w$ or $s$. 
(2) \textbf{Stability:} The performance curves are relatively flat within a reasonable range (e.g., $2 \le w \le 6$), indicating that the method is not overly sensitive to precise tuning. 
(3) \textbf{Robustness:} The same optimal configurations ($w=4, s=1$) yield peak performance across both ID and OOD scenarios, confirming its strong generalization. 

Specifically, while window size $w=6$ shows marginal gains in ID metrics, $w=4$ provides the best trade-off between performance and computational overhead. Similarly, $s=1$ demonstrates superior generalizability across OOD conditions, effectively preventing the model from collapsing into narrow reward-specific local optima by repeatedly optimizing critical high-SNR timesteps.

\textbf{Summary on Generalization.} In conclusion, these cross-dataset evaluations and sensitivity analyses confirm that MixGRPO's performance gains are not tied to specific hyperparameter artifacts. The consistent superiority of our method over baselines across various configurations and unseen reward models underscores its methodological robustness and strong generalization capabilities.

\begin{table*}[!t]
  \centering
  \caption{Comparison for different moving strategies in cross-dataset experiments. The optimal setting is highlighted in green.}
  \resizebox{0.9\linewidth}{!}{
    \begin{tabular}{l c | c c c | c c c | c c c | c c c}
\hline 
\multirow{3}{*}{\textbf{Strategy}} & \multirow{3}{*}{\textbf{\begin{tabular}[c]{@{}c@{}}Interval\\ Schedule\end{tabular}}} & \multicolumn{6}{c|}{\textbf{Cross-dataset Experiment 1}} & \multicolumn{6}{c}{\textbf{Cross-dataset Experiment 2}} \\
\cline{3-14}
& & \multicolumn{3}{c|}{\textbf{In Domain Dataset (HPD v2)}} & \multicolumn{3}{c|}{\textbf{Out-of-Domain Dataset (Pick-a-Pi v1)}} & \multicolumn{3}{c|}{\textbf{In Domain Dataset (Pick-a-Pi v1)}} & \multicolumn{3}{c}{\textbf{Out-of-Domain Dataset (HPD v2)}} \\
\cline{3-14}
& & \textbf{HPS-v2.1} & \textbf{Pick Score} & \textbf{ImageReward} & \textbf{HPS-v2.1} & \textbf{Pick Score} & \textbf{ImageReward} & \textbf{HPS-v2.1} & \textbf{Pick Score} & \textbf{ImageReward} & \textbf{HPS-v2.1} & \textbf{Pick Score} & \textbf{ImageReward} \\
\hline 
Frozen & / & 0.354 & 0.234 & 1.580 & 0.352 & 0.226 & 1.539 & 0.346 & 0.230 & 1.601 & 0.351 & 0.231 & 1.587 \\
\hline 
Random & Constant & 0.365 & 0.237 & 1.513 & 0.361 & 0.230 & 1.512 & 0.349 & 0.227 & 1.524 & 0.343 & 0.225 & 1.530 \\
\hline 
\multirow{3}{*}{Progressive} & Decay(Linear) & 0.365 & 0.235 & 1.566 & 0.363 & 0.229 & 1.572 & 0.363 & 0.231 & 1.614 & 0.355 & 0.230 & 1.597 \\
& Decay(Exp) & 0.360 & \textbf{0.239} & \textbf{1.632} & 0.364 & 0.232 & 1.612 & 0.361 & 0.234 & \textbf{1.628} & 0.358 & 0.230 & 1.584 \\
\rowcolor{green!15}
& Constant & \textbf{0.367} & 0.237 & 1.629 & \textbf{0.366} & \textbf{0.234} & \textbf{1.622} & \textbf{0.365} & \textbf{0.238} & 1.618 & \textbf{0.359} & \textbf{0.232} & \textbf{1.601} \\
\hline 
\end{tabular}
  }
  \label{tab:ood_strategy}
\end{table*}

\begin{table*}[!t]
  \centering
  \caption{Comparison for different shift intervals $\tau$ in cross-dataset experiments. The optimal setting is highlighted in green.}
  \resizebox{0.9\linewidth}{!}{
    \begin{tabular}{c | c c c | c c c | c c c | c c c}
\hline 
\multirow{3}{*}{\textbf{$\tau$}} & \multicolumn{6}{c|}{\textbf{Cross-dataset Experiment 1}} & \multicolumn{6}{c}{\textbf{Cross-dataset Experiment 2}} \\
\cline{2-13}
& \multicolumn{3}{c|}{\textbf{In Domain Dataset (HPD v2)}} & \multicolumn{3}{c|}{\textbf{Out-of-Domain Dataset (Pick-a-Pi v1)}} & \multicolumn{3}{c|}{\textbf{In Domain Dataset (Pick-a-Pi v1)}} & \multicolumn{3}{c}{\textbf{Out-of-Domain Dataset (HPD v2)}} \\
\cline{2-13}
& \textbf{HPS-v2.1} & \textbf{Pick Score} & \textbf{ImageReward} & \textbf{HPS-v2.1} & \textbf{Pick Score} & \textbf{ImageReward} & \textbf{HPS-v2.1} & \textbf{Pick Score} & \textbf{ImageReward} & \textbf{HPS-v2.1} & \textbf{Pick Score} & \textbf{ImageReward} \\
\hline
15 & 0.366 & 0.237 & 1.509 & 0.359 & 0.227 & 1.470 & 0.358 & 0.234 & 1.610 & 0.353 & 0.228 & 1.542 \\
20 & 0.366 & \textbf{0.238} & 1.610 & 0.360 & 0.228 & 1.619 & 0.361 & 0.237 & 1.615 & 0.357 & \textbf{0.233} & 1.588 \\
\rowcolor{green!15} 
25 & \textbf{0.367} & 0.237 & \textbf{1.629} & \textbf{0.366} & \textbf{0.234} & \textbf{1.623} & \textbf{0.365} & \textbf{0.238} & \textbf{1.618} & \textbf{0.359} & 0.232 & \textbf{1.601} \\
30 & 0.350 & 0.229 & 1.589 & 0.348 & 0.221 & 1.585 & 0.355 & 0.234 & 1.609 & 0.351 & 0.229 & 1.509 \\
\hline 
\end{tabular}
  }
  \label{tab:ood_iterations}
\end{table*}

\begin{table*}[!t]
  \centering
  \caption{Comparison for different window sizes $w$ in cross-dataset experiments. The optimal setting is highlighted in green.}
  \resizebox{0.9\linewidth}{!}{
    \begin{tabular}{c | c c c | c c c | c c c | c c c}
\hline 
\multirow{3}{*}{\textbf{$w$}} & \multicolumn{6}{c|}{\textbf{Cross-dataset Experiment 1}} & \multicolumn{6}{c}{\textbf{Cross-dataset Experiment 2}} \\
\cline{2-13}
& \multicolumn{3}{c|}{\textbf{In Domain Dataset (HPD v2)}} & \multicolumn{3}{c|}{\textbf{Out-of-Domain Dataset (Pick-a-Pi v1)}} & \multicolumn{3}{c|}{\textbf{In Domain Dataset (Pick-a-Pi v1)}} & \multicolumn{3}{c}{\textbf{Out-of-Domain Dataset (HPD v2)}} \\
\cline{2-13}
& \textbf{HPS-v2.1} & \textbf{Pick Score} & \textbf{ImageReward} & \textbf{HPS-v2.1} & \textbf{Pick Score} & \textbf{ImageReward} & \textbf{HPS-v2.1} & \textbf{Pick Score} & \textbf{ImageReward} & \textbf{HPS-v2.1} & \textbf{Pick Score} & \textbf{ImageReward} \\
\hline 
1 & 0.359 & 0.232 & 1.571 & 0.349 & 0.223 & 1.445 & 0.349 & 0.229 & 1.553 & 0.346 & 0.221 & 1.565 \\
2 & 0.366 & 0.235 & 1.618 & 0.363 & 0.230 & 1.614 & 0.363 & 0.234 & 1.597 & 0.359 & 0.230 & 1.585 \\

\rowcolor{green!15} 
4 & 0.367 & 0.237 & \textbf{1.629} & \textbf{0.366} & \textbf{0.234} & \textbf{1.623} & 0.365 & \textbf{0.238} & \textbf{1.618} & 0.359 & \textbf{0.232} & \textbf{1.601} \\
6 & \textbf{0.370} & \textbf{0.238} & 1.624 & 0.364 & 0.232 & 1.620 & \textbf{0.366} & 0.237 & 1.608 & \textbf{0.362} & 0.232 & 1.594 \\
\hline 
\end{tabular}
  }
  \label{tab:ood_size}
\end{table*}

\begin{table*}[!t]
  \centering
  \caption{Comparison for different window strides $s$ in cross-dataset experiments. The optimal setting is highlighted in green.}
  \resizebox{0.9\linewidth}{!}{
    \begin{tabular}{c | c c c | c c c | c c c | c c c}
\hline 
\multirow{3}{*}{\textbf{$s$}} & \multicolumn{6}{c|}{\textbf{Cross-dataset Experiment 1}} & \multicolumn{6}{c}{\textbf{Cross-dataset Experiment 2}} \\
\cline{2-13}
& \multicolumn{3}{c|}{\textbf{In Domain Dataset (HPD v2)}} & \multicolumn{3}{c|}{\textbf{Out-of-Domain Dataset (Pick-a-Pi v1)}} & \multicolumn{3}{c|}{\textbf{In Domain Dataset (Pick-a-Pi v1)}} & \multicolumn{3}{c}{\textbf{Out-of-Domain Dataset (HPD v2)}} \\
\cline{2-13}
& \textbf{HPS-v2.1} & \textbf{Pick Score} & \textbf{ImageReward} & \textbf{HPS-v2.1} & \textbf{Pick Score} & \textbf{ImageReward} & \textbf{HPS-v2.1} & \textbf{Pick Score} & \textbf{ImageReward} & \textbf{HPS-v2.1} & \textbf{Pick Score} & \textbf{ImageReward} \\
\hline 
\rowcolor{green!15} 
1 & 0.367 & 0.237 & \textbf{1.629} & \textbf{0.366} & \textbf{0.234} & \textbf{1.623} & 0.364 & \textbf{0.238} & \textbf{1.618} & \textbf{0.359} & \textbf{0.232} & \textbf{1.601} \\
2 & 0.357 & 0.236 & 1.575 & 0.357 & 0.233 & 1.587 & 0.363 & 0.234 & 1.574 & 0.354 & 0.231 & 1.591 \\
3 & \textbf{0.370} & 0.236 & 1.578 & 0.364 & 0.230 & 1.579 & \textbf{0.368} & 0.237 & 1.586 & 0.358 & 0.228 & 1.585 \\
4 & 0.368 & \textbf{0.238} & 1.575 & 0.349 & 0.224 & 1.573 & 0.366 & 0.231 & 1.566 & 0.357 & 0.225 & 1.568 \\
\hline 
\end{tabular}
  }
  \label{tab:ood_step}
\end{table*}

\section{Ablation Study on Intra-group Initial Noise}
\label{app:initial_noise}

In this section, we investigate the impact of the initial noise setting within a sampling group during the MixGRPO training process. A potential concern in group-based reinforcement learning is that diverse initial noises across different samples within the same group might introduce unwanted variance, potentially leading to biased reward signals or "reward hacking."

Following the observations in \text{DanceGRPO} \cite{xue2025dancegrpo}, we adopt a strategy of fixing the intra-group initial noise. This approach ensures that the performance variations within a group primarily stem from the model's policy updates rather than the stochasticity of the starting latent states. By neutralizing the noise factor, the relative advantage of certain generations becomes a more reliable indicator for policy improvement.

To validate this design choice, we conducted an ablation study comparing MixGRPO with and without fixed intra-group initial noise. As shown in Table~\ref{tab:noise_ablation}, fixing the initial noise consistently yields superior results across all preference benchmarks, including HPS-v2.1, Pick Score, and ImageReward. These results demonstrate that stabilizing the initial conditions effectively mitigates reward hacking and leads to more robust policy optimization.

\begin{table}[h]
    \centering
    \caption{Ablation study of intra-group initial noise on MixGRPO performance.}
    \label{tab:noise_ablation}
    \small
    \begin{tabular}{lcccc}
        \toprule
        \textbf{Method} & \textbf{HPS-v2.1 $\uparrow$} & \textbf{Pick Score $\uparrow$} & \textbf{ImageReward $\uparrow$} \\
        \midrule
        MixGRPO (w/o fixed initial noise) & 0.342 & 0.228 & 1.448 \\
        MixGRPO (w/ fixed initial noise) & \textbf{0.367} & \textbf{0.237} & \textbf{1.629} \\
        \bottomrule
    \end{tabular}
\end{table}

\section{Hybrid Inference for Solving Reward Hacking}
\label{app: mixsampling_for_hacking}
As discussed in Section~\ref{sec: limitation}, reward hacking stems from the limited evaluation capabilities of the reward model. To address reward hacking and improve visualization, we employ the hybrid inference strategy from~\cite{flow_grpo_issue7}, which uses the post-trained model for low-SNR (signal-to-noise ratio) steps and the original model for high-SNR steps during inference-time sampling. In our experiments, we also applied hybrid inference to the other baseline models to ensure a fair and consistent comparison.

\begin{figure}[H]
    \centering
    \includegraphics[width=\linewidth]{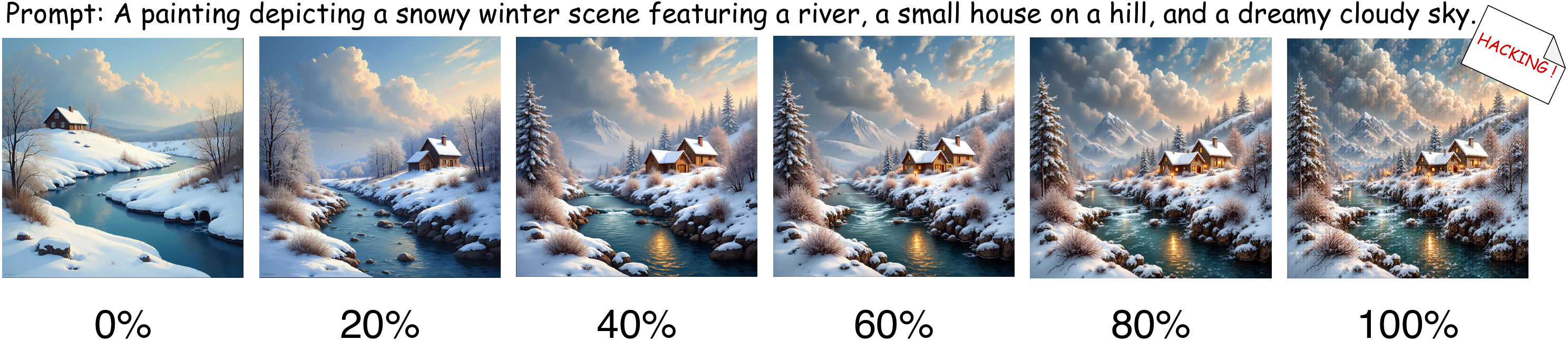}
    \caption{Qualitative comparison with  different hybrid inference percentages. We found 80\% to be the optimal value, as it maximizes image quality without causing over-optimization.}
    \label{fig: mix_sampling_hacking}
\end{figure}

We employ hybrid inference and introduce the hybrid percent \( p_{\text{mix}} \). This means that the initial, high-SNR \( p_{\text{mix}}T\) denoising steps, are handled by the model trained with GRPO, while the remaining denoising process is finished by the original model. Table~\ref{tab: mix_sampling} and Figure~\ref{fig: mix_sampling_hacking} respectively illustrate the changes in performance and images as \( p_{\text{mix}} \) increases under the multi-rewards training scenario. The experimental results demonstrate that \( p_{\text{mix}} = 80\% \) is an optimal empirical value that effectively mitigates hacking while maximizing alignment with human preferences. 
\begin{table}[ht]
\centering
\caption{Comparison with different hybrid inference percentages}
\resizebox{0.8\linewidth}{!}{
\begin{tabular}{lcccc}
\toprule
\textbf{$p_{\text{mix}}$} & \textbf{HPS-v2.1} & \textbf{Pick Score} & \textbf{ImageReward} & \textbf{Unified Reward} \\
\midrule
0\%   & 0.313 & 0.226 & 1.089 & 3.369 \\
20\%  & 0.342 & 0.233 & 1.372 & 3.386 \\
40\%  & 0.356 & 0.235 & 1.539 & 3.395 \\
60\%  & 0.362 & 0.236 & 1.598 & 3.407 \\
80\%  & 0.366 & \textbf{0.238} & \textbf{1.610} & \textbf{3.411} \\
100\% & \textbf{0.369} & 0.238 & 1.607 & 3.378 \\
\bottomrule
\end{tabular}
}
\label{tab: mix_sampling}
\end{table}

\section{Coefficients-Preserving Sampling}
\label{app: CPS}
In our \name{} framework, introducing stochasticity during the inference phase is crucial for effective exploration in reinforcement learning. While a common practice involves the use of Stochastic Differential Equations (SDEs), we adopt Coefficients-Preserving Sampling (CPS)~\cite{wang2025coefficients} as a more refined alternative to maintain the integrity of the probability path.

The standard SDE-based sampling, often discretized through the Euler-Maruyama method, follows the update rule:
\begin{equation}
    \mathbf{x}_{t_{i-1}} = \mathbf{x}_{t_i} - \mathbf{v}_{t_i}(\mathbf{x}_{t_i}, t_i) \Delta t + \sqrt{2\sigma^2 \Delta t} \mathbf{\epsilon}_i, \quad \mathbf{\epsilon}_i \sim \mathcal{N}(\mathbf{0}, \mathbf{I})
\end{equation}
Although this formulation provides the necessary stochasticity, it tends to inject excessive independent noise at each discretization step, resulting in "grainy" artifacts in the generated samples. As observed in our preliminary experiments, such high-frequency noise can lead to reward hacking, where reward models (e.g., Pick Score~\cite{kirstain2023pick} or HPS-v2.1~\cite{wu2023human}) prioritize low-level textures over structural coherence, thereby hindering the convergence of RL optimization.

\begin{figure*}[ht]
    \centering
    \includegraphics[width=\linewidth]{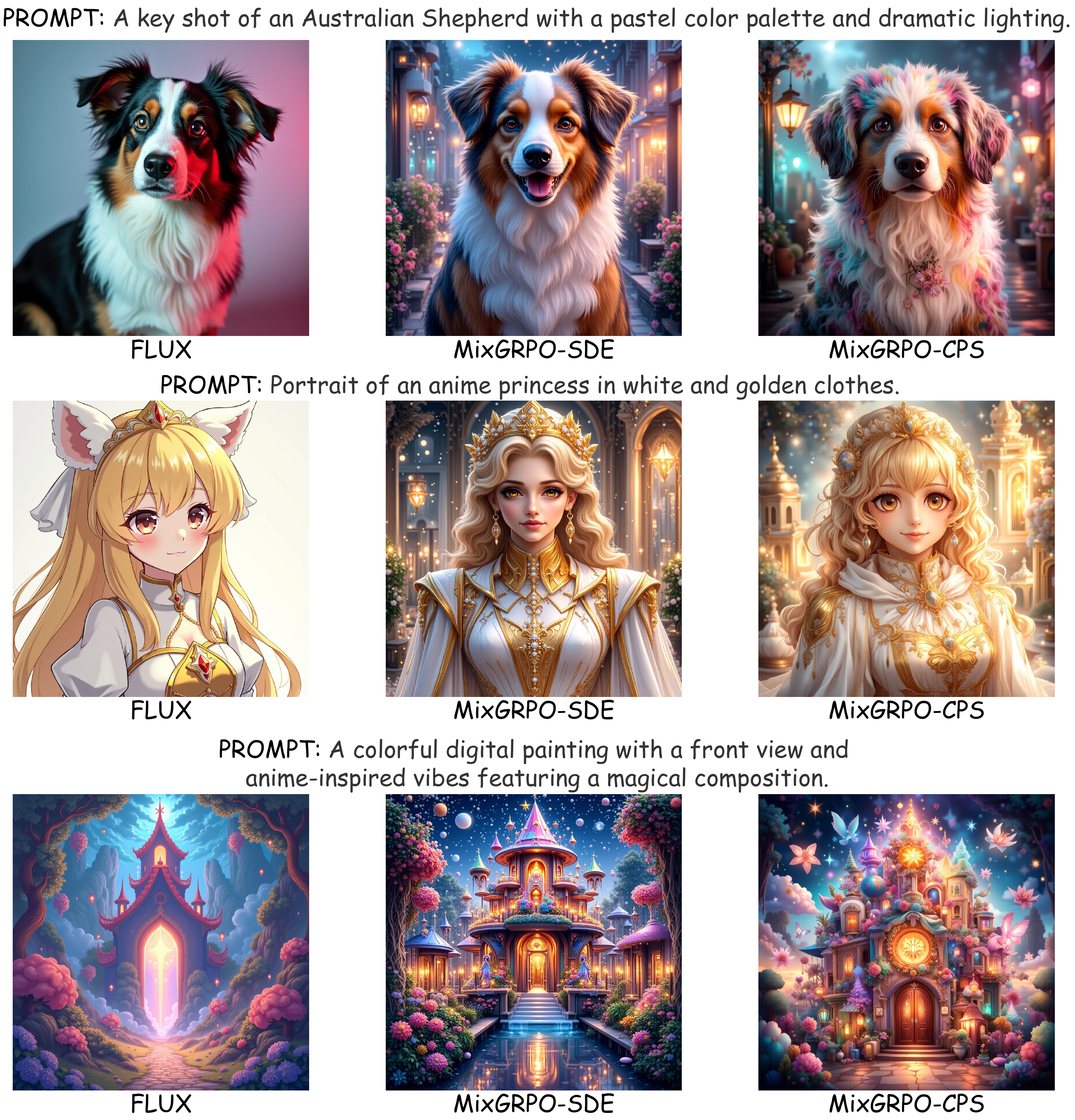}
    \caption{Comparison of the visualization results of FLUX, SDE sampling and CPS sampling}
    \label{fig:cps_visual}
\end{figure*}

To address these issues, we utilize CPS~\cite{wang2025coefficients}, which reformulates the transition process by drawing inspiration from DDIM~\cite{song2020denoising} framework. The update rule for CPS~\cite{wang2025coefficients} is defined as:
\begin{equation}
    \mathbf{x}_{t_{i-1}} = \frac{1-t_{i-1}}{1-t_i} \mathbf{x}_{t_i} + \left( t_{i-1} - \frac{1-t_{i-1}}{1-t_i} t_i \right) \mathbf{v}_{t_i} + \sigma_{t_i} \mathbf{\epsilon}_i
\end{equation}
The core advantage of CPS~\cite{wang2025coefficients} lies in its strategic construction of coefficients for $\mathbf{x}_{t_i}$ and $\mathbf{v}_{t_i}$, which preserves the linear interpolation structure of Flow Matching while introducing controlled stochasticity via $\sigma_{t_i}$. By effectively eliminating sampling artifacts inherent in SDE-based methods, CPS~\cite{wang2025coefficients} yields cleaner images that provide more reliable feedback for \name{}. In our implementation, we set $NFE_{\pi_{\theta_{old}}} = 25$, window size $w = 4$, and window stride $s = 1$ with fixed initial noise, employing HPS-v2.1~\cite{wu2023human}, Pick Score~\cite{kirstain2023pick}, and ImageReward~\cite{xu2023imagereward} as multi-reward metrics. Both the original SDE sampling and CPS~\cite{wang2025coefficients} sampling were trained for 300 steps and evaluated on the HPDv2 dataset~\cite{wu2023human}. The quantitative comparison is moved to the main ablation section (Table~\ref{tab: sde_cps}), and qualitative comparisons are shown in Figure~\ref{fig:cps_visual}.

\section{More Visualized Results}
\label{app: more_visualized_results}

\begin{figure*}[ht]
    \centering
    \includegraphics[width=\linewidth]{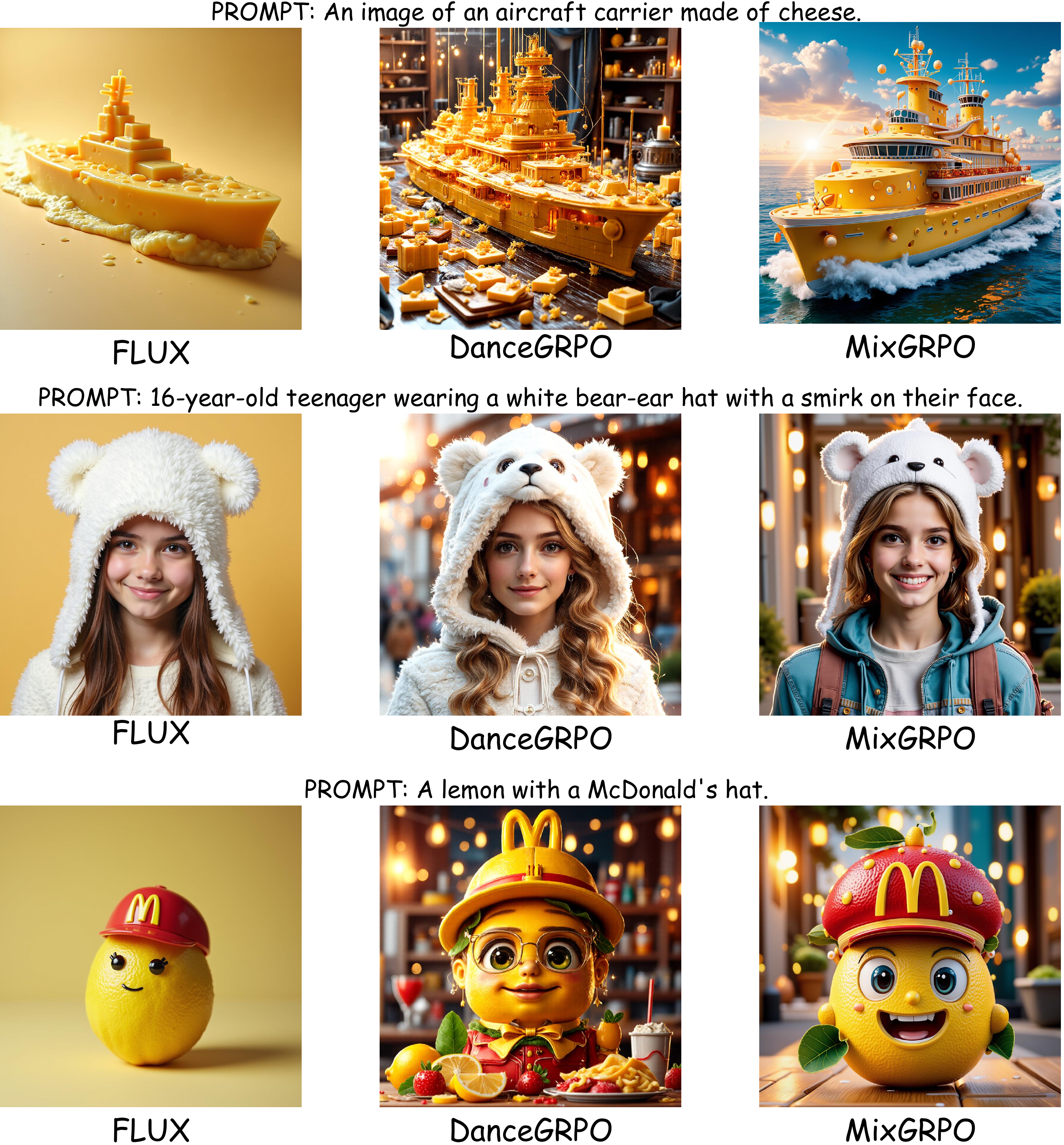}
    \caption{Comparison of the visualization results of FLUX, DanceGRPO, and \name{} under HPS-v2.1 as the reward model.}
    \label{fig:hps_only}
\end{figure*}

\begin{figure*}[ht]
    \centering
    \includegraphics[width=\linewidth]{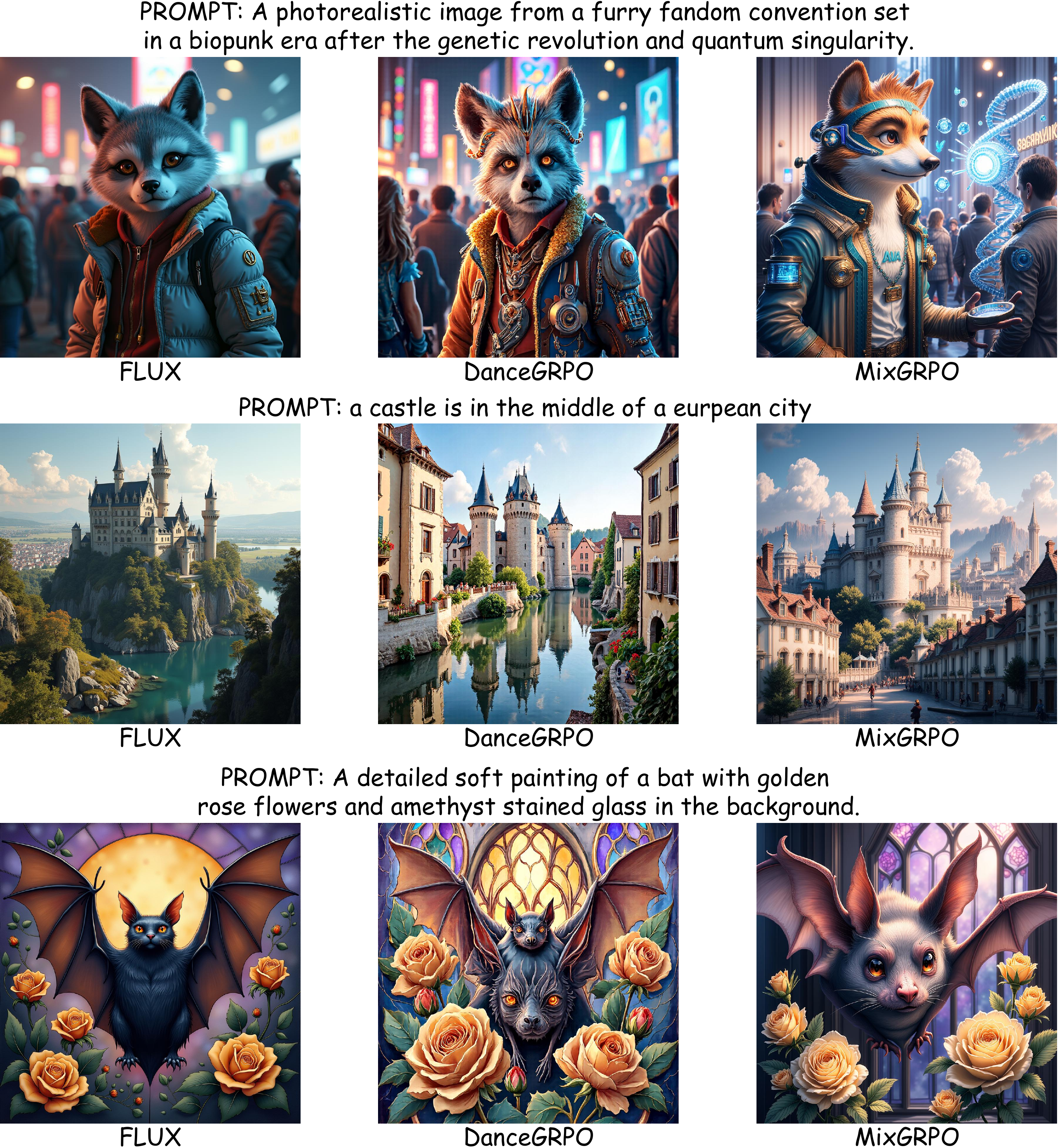}
    \caption{Comparison of the visualization results of FLUX, DanceGRPO, and \name{} under HPS-v2.1 and CLIP Score as multi-reward models.}
    \label{fig:hps_clip}
\end{figure*}

\begin{figure*}[ht]
    \centering
    \includegraphics[width=\linewidth]{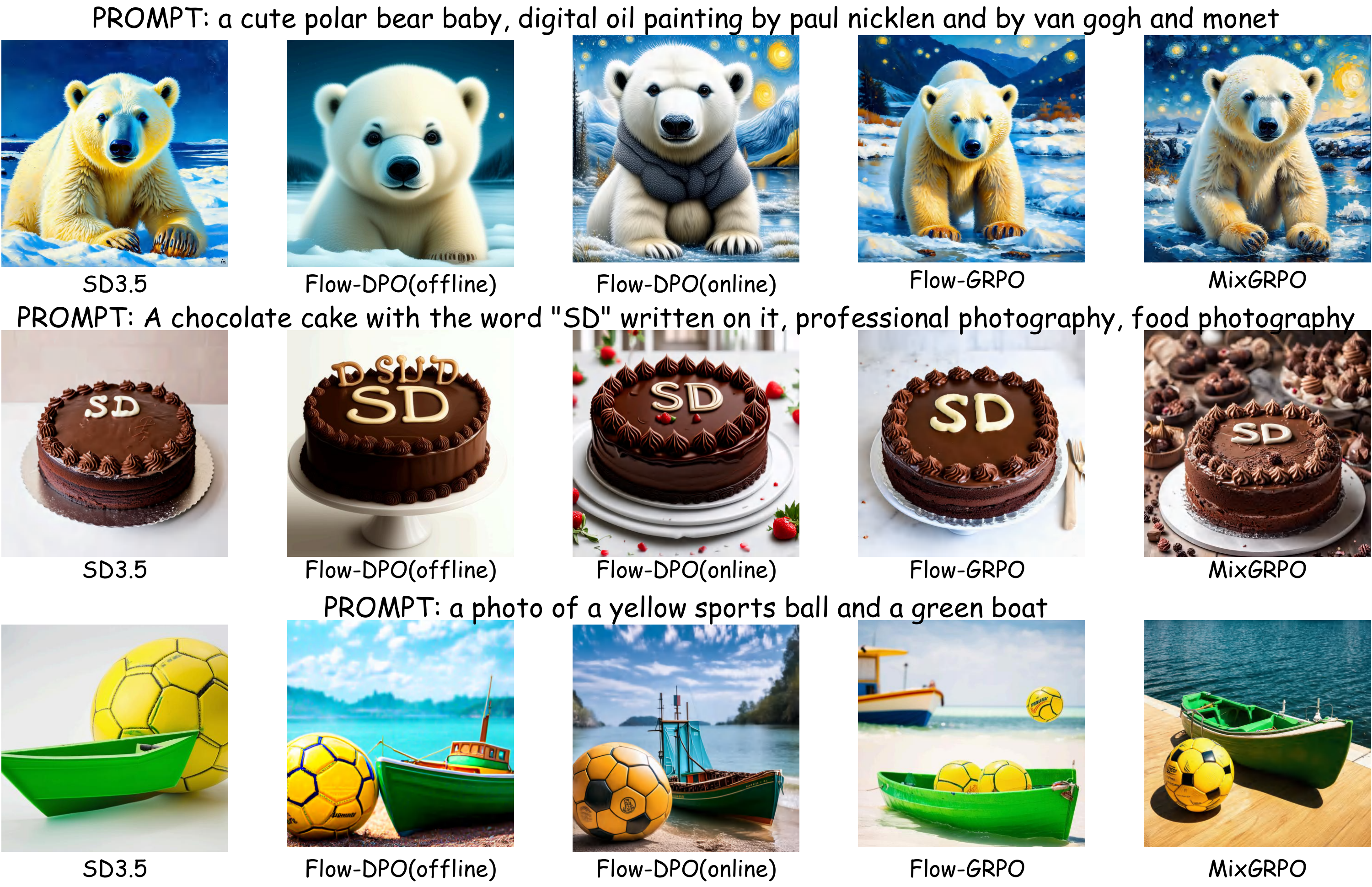}
    \caption{Comparison of the visualization results of SD3.5-M, offline DPO, online DPO, Flow-GRPO and \name{} under HPS-v2.1, Pick Score and ImageReward as multi-reward models.}
    \label{fig:flow_grpo}
\end{figure*}

\clearpage



\end{document}